\documentclass[11pt]{article}

\usepackage[final]{acl}

\usepackage{times}
\usepackage{latexsym}
\usepackage{tikz}
\usetikzlibrary{arrows.meta, positioning}
\usepackage{float} 
\usepackage{hyperref}
\usepackage[T1]{fontenc}

\usepackage[utf8]{inputenc}

\usepackage{microtype}

\usepackage{inconsolata}

\usepackage{graphicx}
\usepackage{multirow}
\usepackage{tablefootnote}
\usepackage{adjustbox}
\usepackage{booktabs}
\usepackage{nicematrix} 

\newcommand{\uoslogo}{\raisebox{3.2pt}{\includegraphics[scale=0.065]{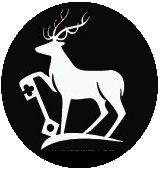}}}
\newcommand{\PAIlogo}{\raisebox{3.4pt}{\includegraphics[scale=0.08]{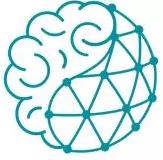}}}

\usepackage{microtype}
\usepackage{hyperref}
\usepackage{url}
\usepackage{booktabs}

\usepackage{xspace}
\usepackage{multirow}   
\usepackage{graphicx} 

\usepackage{tikz}
\usetikzlibrary{arrows.meta, positioning, shapes.geometric}

\makeatother



\newcommand{\alope}{\texttt{ALOPE}\xspace}
\newcommand{\healthcare}{\texttt{Healthcare}\xspace}
\newcommand{\legal}{\texttt{Legal}\xspace}
\newcommand{\tourism}{\texttt{Tourism}\xspace}
\newcommand{\general}{\texttt{General}\xspace}

\title{Domain-Specific Quality Estimation for Machine Translation in Low-Resource Scenarios}

\author{
     Namrata Patil Gurav\uoslogo, Akashdeep Ranu\uoslogo, Archchana Sindhujan\PAIlogo, Diptesh Kanojia\PAIlogo
    \\ [.35em]
    \PAIlogo Institute for People-Centred AI,  \\ \uoslogo University of Surrey, United Kingdom\\ [.15em]
    \texttt{\{np00996, ar02258, a.sindhujan, d.kanojia\}@surrey.ac.uk}\\ 
}

\begin{document}
\maketitle
\begin{abstract}
Quality Estimation (QE) is essential for assessing machine translation quality in reference-less settings, particularly for domain-specific and low-resource language scenarios. In this paper, we investigate sentence-level QE for English$\rightarrow$Indic machine translation across four domains (\healthcare, \legal, \tourism, and \general) and five language pairs. We systematically compare zero-shot, few-shot, and guideline-anchored prompting across selected closed-weight and open-weight LLMs. Findings indicate that while closed-weight models achieve strong performance \textit{via} prompting alone, prompt-only approaches remain fragile for open-weight models, especially in high-risk domains. To address this, we adopt \alope, a framework for LLM-based QE which uses Low-Rank Adaptation with regression heads attached to selected intermediate Transformer layers. We also extend \alope with the recently proposed Low-Rank Multiplicative Adaptation (LoRMA) for this work.
Our results show that intermediate-layer adaptation consistently improves QE performance, with gains in semantically complex domains, indicating a way ahead for robust QE in practical scenarios. We release code and domain-specific QE datasets publicly for further research\footnote{\href{https://github.com/surrey-nlp/ALOPE/tree/main/Domain-based-QE-with-ALOPE}{https://github.com/surrey-nlp/ALOPE/tree/main/Domain-based-QE-with-ALOPE}}.

\end{abstract}

\section{Introduction}

Quality Estimation (QE) enables the output of Machine Translation (MT) systems to be evaluated at scale without requiring reference translations \citep{zerva-etal-2022-findings}. Unlike traditional MT evaluation metrics such as BLEU or METEOR, QE directly predicts a quality score for a source–translation pair, making it especially suitable for real-world deployment scenarios where reference translations are unavailable. QE can be performed at multiple granularities, and our work focuses on segment-level QE, predicting Direct Assessment (DA) scores on a continuous scale of $(0 \leq x \leq 100)$ for the given translation \citep{graham2013continuous}. The ground-truth DA score is obtained by averaging the scores from three or more human annotators.

While fine-grained annotation frameworks such as Multi-dimensional Quality Metrics (MQM) \citep{burchardt-2013-multidimensional} provide detailed insights into translation errors, they impose a substantial cognitive and temporal burden on annotators. In practice, reliable QE enables translation systems to prioritise human intervention for high-risk outputs, which is particularly important in sensitive or specialised domains \citep{zerva-etal-2024-findings}.
Despite the widespread adoption of neural machine translation, translation quality remains uneven across languages and domains, particularly outside high-resource, \general-domain settings \citep{specia2018quality,zhao-etal-2024-survey}. This disparity is especially pronounced for English$\rightarrow$Indic language pairs \citep{sindhujan-etal-2025-llms}, where rich morphology, frequent code-mixing, script variation, and limited availability of high-quality parallel data continue to pose persistent challenges for both MT and evaluation \citep{zhao-etal-2024-survey}. 

Although MT output for \general content is often fluent, translations in domain-specific settings such as \healthcare, \legal, and \tourism remain fragile, as models are less exposed to specialised terminology and domain-specific constructions during training \citep{specia2018quality}. Even minor errors involving negation, numerical values, units, or specialised terminology can result in substantial meaning changes, with potentially serious real-world consequences, particularly in high-risk domains such as \healthcare and \legal \citep{info16100916}. These limitations underscore the need for robust domain-aware QE mechanisms that can reliably identify problematic translations in the absence of reference translations before deployment \citep{zerva-etal-2024-findings}.

Recent advances in Large Language Models (LLMs) have enabled QE through prompt-based scoring of source–translation pairs, offering an alternative in low-resource and domain-specific settings where supervised QE data are limited \citep{brown2020language}. However, prior work shows that prompt-only LLM-based QE remains inferior to state-of-the-art encoder-based QE models, particularly for sentence-level regression tasks \citep{zerva-etal-2022-findings,info16100916}. A key limitation is that LLMs are optimised for next-token prediction rather than regression-oriented objectives such as predicting DA scores. Prompting or instruction tuning alone does not introduce regression-specific training signals, often resulting in unstable predictions \citep{zhao2021calibrate,sindhujan-etal-2025-alope}. Furthermore, most LLM-based QE approaches rely solely on representations from the final Transformer layer, despite growing evidence that intermediate layers may better encode cross-lingual and semantic alignment for low-resource languages \citep{kargaran2025layerwise}.

Our work addresses these limitations by investigating domain-aware QE for English$\rightarrow$Indic translation language pairs through a dual-track evaluation: (i) systematic comparison of prompt-only approaches across closed-weight and open-weight LLM families, and (ii) lightweight, parameter-efficient fine-tuning based approach where open-weight prompt-only methods prove insufficient. 

Building on the \alope framework \citep{sindhujan-etal-2025-alope}, which demonstrated that intermediate Transformer layers encode more stable QE-relevant signals than final-layer representations, we extend this approach to domain-specific, low-resource English$\rightarrow$Indic settings ($\S$~\ref{subsec:ALOPE}). \alope attach regression heads to informative intermediate layers and update only a minimal parameter subset using Low-Rank Adaptors \citep{hu2022lora}, maintaining computational efficiency while improving the QE performance where deployment constraints preclude closed-weight API access.

We conduct a comprehensive evaluation across four domains (\healthcare, \legal, \tourism, and \general) and five Indic languages (Hindi, Marathi, Tamil, Telugu, and Gujarati). Our experiments systematically compare prompt-only baselines (zero-shot, few-shot, and guideline-anchored prompting) and \alope, analysing performance differences between closed-weight models and open-weight models. Critically, our study establishes when lightweight \alope-based methods provide value versus when strong prompting alone suffices, offering practical deployment guidance for resource-constrained QE scenarios. The \textbf{main contributions} of this work are as follows:
\begin{itemize}
\item We provide a rigorous comparison of prompt-only QE strategies across closed-weight and open-weight LLM families, revealing that closed-weight models with guideline-anchored prompting achieve robust performance within domains and language pairs.

\item We demonstrate that a lightweight \alope-inspired approach, which leverages selected intermediate Transformer layer representations, achieves competitive QE performance in resource-constrained settings. Across the majority of the domains and language pairs, intermediate Transformer layers consistently yield stronger QE signals than final-layer representations.

\item  We establish a simple practical framework for QE deployment in low-resource, domain-sensitive settings, providing clear guidance on when to prioritise strong prompting versus when to apply lightweight adapter-based methods.

\end{itemize}

\section{Background}
Machine Translation (MT) quality remains uneven across language pairs and domains despite advances in neural and Transformer-based architectures \citep{specia2018quality,zhao-etal-2024-survey}. This disparity is pronounced for English$\rightarrow$Indic translation, where rich morphology, code-mixing, script diversity, and limited parallel data pose persistent challenges  ~\citep{info16100916}. Domain-specific translation in high-risk contexts (Healthcare, Legal) is particularly fragile, as publicly available corpora are skewed toward general web and news content, leaving specialised terminology and discourse structures under-represented  ~\citep{specia2018quality,zhao-etal-2024-survey}. Recent domain-focused evaluations further show that MT quality is strongly domain dependent: systems that perform well on General text often degrade substantially in specialised domains such as Healthcare, Legal, Literary, or User-generated content  ~\citep{makela2024openwho,gupta-etal-2024-domain}. In the Healthcare domain, LLMs leveraging document-level context have been shown to outperform traditional neural MT systems, while this advantage largely disappears in General or News translation, highlighting the interaction between domain characteristics and model effectiveness  ~\citep{makela2024openwho}.

Quality Estimation (QE) addresses this by predicting translation quality without reference translations, enabling scalable assessment ~\citep{zerva-etal-2022-findings}. Segment-level QE using Direct Assessment (DA) scores provides a practical alternative to fine-grained frameworks like MQM, balancing interpretability with annotation efficiency \citep{graham2013continuous}. However, QE performance in low-resource, domain-shifted settings remains constrained by limited labelled data and high-impact error types \citep{zhao-etal-2024-survey}, which are particularly prevalent in specialised domains where terminology misuse or semantic distortion can be critical.

LLMs offer prompt-based, reference-free QE but suffer from score compression, inconsistent calibration, and weak sensitivity to subtle errors due to optimisation for next-token prediction rather than regression objectives \citep{zerva-etal-2024-findings,kocmi-federmann-2023-gemba}. Evidence from domain adaptation studies suggests that while fine-tuning LLM-based MT models on in-domain data consistently improves translation quality, the effectiveness of adaptation depends on the availability and diversity of domain-specific data \citep{patel-etal-2024-businessit}. In the \legal domain, fine-tuning a strong multilingual pre-trained model has been shown to outperform models explicitly pre-trained for the target domain, indicating that robust \general-purpose representations combined with targeted adaptation are often more effective than domain-specific pre-training alone \citep{singh-etal-2025-svnit}. When in-domain data is scarce or rapid adaptation is required, in-context learning provides a viable alternative: selecting few-shot examples based on topic similarity can significantly improve translation quality for unseen domains, though gains depend on balancing relevance with sufficient example diversity \citep{li-etal-2024-topicguided}.

Cross-lingual QE signals are not uniformly distributed across Transformer layers, intermediate layers encode more stable semantic and alignment information than final layers, especially for low-resource languages \citep{kargaran2025layerwise,tenney2019bert}. Parameter-efficient methods such as Low-Rank Adaptation (LoRA) enable task-specific fine-tuning by updating minimal parameters while keeping base models frozen \citep{hu2022lora,dettmers2023qlora}. The \alope framework ($\S$~\ref{subsec:ALOPE}) extends this by attaching lightweight regression heads to intermediate layers, targeting QE-relevant representations \citep{sindhujan-etal-2025-alope}. Our work investigates whether intermediate-layer adaptation provides consistent benefits across domains and low-resource English$\rightarrow$Indic pairs, or whether domain characteristics and pre-training coverage moderate its effectiveness.

\begin{table}[t]
\centering
\small
\setlength{\tabcolsep}{6pt}
\resizebox{\columnwidth}{!}{
\begin{tabular}{llrrr}
\toprule
\textbf{Domain} & \textbf{Langs} &
\textbf{Train} & \textbf{Test} \\
\midrule
\healthcare &
Hi/Mr/Ta/Gu &
13{,}280 & 1{,}660 \\

\legal &
Gu/Ta/Te &
6{,}160 & 770 \\

\tourism &
Hi/Mr/Te &
13{,}840 & 1{,}730 \\

\general &
Hi/Mr/Ta/Te/Gu &
18{,}880 & 2{,}360 \\

\bottomrule
\end{tabular}
}
\caption{Indic-Domain-QE dataset composition, showing domain coverage, language pairs, and the number of instances in the train and test splits aggregated across languages.}
\label{tab:qe_datasets_merged}
\end{table}

\section{Methodology}
\begin{figure*}[t]
\centering
\includegraphics[width=0.95\textwidth,trim={0 0 7.1cm 0},clip]{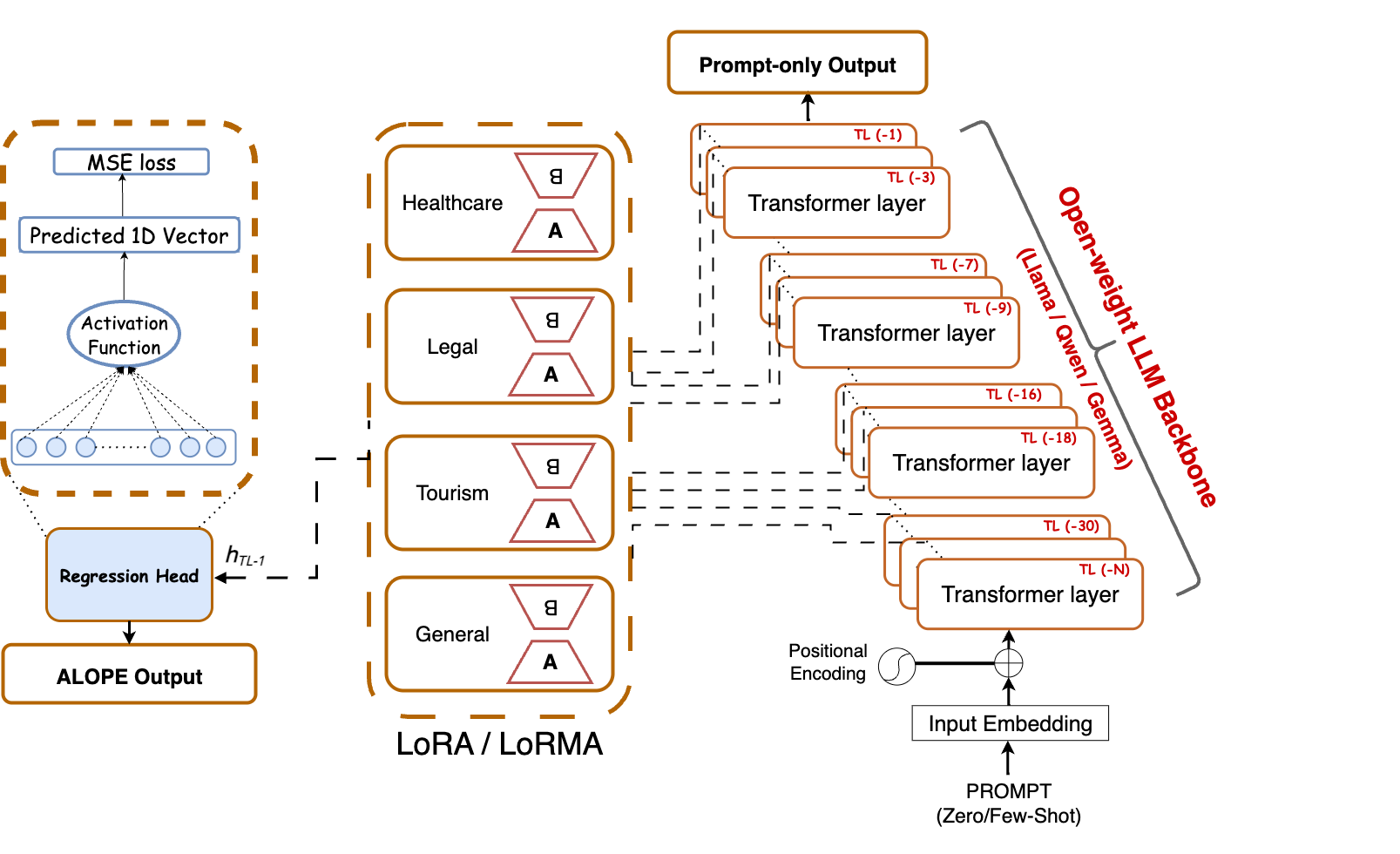}
\caption{Methodological Framework uses open-weight models for (i) prompt-only approaches (zero-shot, few-shot, guideline-anchored), and (ii) \alope adaptation with LoRA/LoRMA.}
\label{fig:research_framework}
\end{figure*}

The methodology adopted in this research provides a systematic and reproducible approach to investigating domain-specific QE for English–Indic language pairs using LLMs. Unlike generic MT evaluation, this research emphasises four practical domains where translation errors can have tangible effects on daily life. Starting from a structured dataset with human-annotated DA scores across four domains and five language pairs, we evaluate QE approaches along two parallel tracks: Prompt-Only approaches and the ALOPE-based approach. Within Prompt-only approaches, we evaluate closed-weight models for comparison, whereas open-weight models are used as backbone within both approaches. Figure~\ref{fig:research_framework} provides an overview of our methodological framework using open-weight models. 

\subsection{Dataset Construction}

The Indic-Domain-QE dataset was created to support systematic evaluation of QE in domain-sensitive contexts. Texts were sourced from publicly available bilingual resources and curated domain-specific materials. Domain characterisation follows standard MT and QE research practices \citep{specia2018quality,zhao-etal-2024-survey}, based on provenance and communicative function of source texts.

\healthcare data consist of patient-facing medical content such as information leaflets, public health advisories, and community health articles, exhibiting high terminology density and sensitivity to negation and numerical expressions. \legal texts are drawn from contracts, policy excerpts, and official notices, characterised by formal register, modality, and scope-defining constructions. \tourism data originate from brochures, attraction descriptions, and travel guidelines, rich in named entities and culturally grounded references. \general-domain data comprise broad-coverage sentences from encyclopaedic and public-information sources, providing a baseline with minimal domain-specific constraints.

Each entry contains an English source, its translation, a domain label, and human-annotated DA scores. Annotators were trained following established DA protocols \citep{graham2013continuous}. Table~\ref{tab:qe_datasets_merged} provides a detailed breakdown of the instances used in this study, with the training split employed for fine-tuning \alope and the test split used for evaluation under both prompt-only and \alope-based settings.

\subsection{Prompt-only Approaches}
We evaluate three prompting strategies for QE, while keeping the task instruction fixed across domains and language pairs, varying only the presence of in-context examples and explicit guidance. Selected closed-weight and open-weight models are evaluated under zero-shot prompting, few-shot prompting without guidelines, and few-shot prompting with guideline anchoring. 

\paragraph{Zero-shot.} The model receives only a natural-language task instruction and the input sentence pair, without in-context demonstrations (Prompt: App.~\ref{app:prompts}-Figure ~\ref{fig:zero-shot}). This setup relies entirely on knowledge from pre-training of the selected model\citep{brown2020language}. While simple and cost-effective, zero-shot prompting often leads to score compression and unstable calibration in regression tasks.

\paragraph{Few-shot (without guidelines).} This includes $1$ to $5$ labelled input-output examples in the prompt, with minimal additional instruction (Prompt: App.~\ref{app:prompts} - Figure ~\ref{fig:fewshot_fewshot_without_guideline}). This in-context learning setup conditions the model on representative examples without parameter updates \citep{brown2020language}. Prior work shows few-shot exemplars improve task adherence but may suffer from variability when explicit scoring criteria are absent.

\paragraph{Few-shot (with guidelines).} This augments the exemplar-based setup with an explicit scoring rubric defining the intended semantics of the output scale (Prompt: App.~\ref{app:prompts}-Figure ~\ref{fig:fewshot_with_guideline}). Guidelines clarify numerical score assignment, reducing ambiguity. Prior studies demonstrate that such explicit constraints improve output consistency and reduce prompt sensitivity \citep{mishra2022cross,zhao2021calibrate}.

Prompt-only evaluations are conducted with two types of models:
\begin{itemize}
    \item \textbf{Closed-weight models:} Gemini-1.5-Pro and Gemini-2.5-Pro, accessed via API, serving as strong prompt-only baselines without parameter updates.
    
    \item \textbf{Open-weight models:} LLaMA-3.2-3B Instruct, LLaMA-3.1-8B Instruct, Qwen3-14B Instruct, and Gemma-3-27B Instruct.  
\end{itemize}

\subsection{ALOPE Framework}\label{subsec:ALOPE}

\alope (\textbf{A}daptive \textbf{L}ayer \textbf{OP}timization for Translation Quality \textbf{E}stimation) \citep{sindhujan-etal-2025-alope} is a parameter-efficient fine-tuning-based framework that attaches regression heads to selected intermediate Transformer layers and updates only a minimal parameter subset using LoRA \citep{hu2022lora}. The original framework explores both single-layer heads and multi-layer variants with dynamic weighting. We only adopt the simplified single-layer configuration for feasibility and reproducibility, evaluating both LoRA and LoRMA adapter variants within the same framework. LoRMA stands for Low-Rank Multiplicative Adaptation, and unlike LoRA, which fine-tunes models by additively injecting low-rank weight updates, LoRMA adapts models by multiplicatively modulating existing weights~\citep{bihany-etal-2025-lorma}.

We conduct all \alope experiments using LLaMA-3.2-3B Instruct as the backbone model, following the best-performing configuration reported in the original \alope study \citep{sindhujan-etal-2025-alope}. We explore rank configurations $R \in \{32, 64, 128\}$ with scaling factor $\alpha \in \{16, 32\}$, extracting representations from Layers $\{-1, -7, -9, -11\}$. The rank $R$ determines the capacity of the low-rank decomposition used to parameterise the weight update, with each update factorised into two trainable matrices while keeping the pre-trained weights frozen. The scaling factor $\alpha$ rescales the update by $\alpha/R$, stabilising training and enabling effective adaptation across different rank settings without extensive hyperparameter tuning \citep{hu2022lora}.

The regression head is a lightweight two-layer feed-forward network with ReLU activation, mapping layer representations to scalar DA predictions. Training uses mean squared error (MSE) loss on gold DA scores. All the adapter-based experiments use 4-bit quantized base models via QLoRA (Quantized-LoRA) to ensure computational efficiency~\citep{dettmers2023qlora}.

\paragraph{Evaluation Metrics.} We evaluate model performance using two correlation-based metrics. Spearman’s rank correlation ($\rho$)~\citep{sedgwick2014spearman} measures the agreement between the relative ordering of predicted and gold Direct Assessment (averaged across annotators) scores, making it well-suited for Quality Estimation where reliable ranking of translations is often more critical than exact score values. Owing to its robustness to scale differences and outliers, $\rho$ is used as our primary evaluation metric. Pearson’s correlation ($r$)~\citep{cohen2009pearson} quantifies linear agreement between predicted and gold DA scores, indicating how closely model outputs match human scores on the same numeric scale. We report $r$ as the additional metric. Both metrics are computed per language pair and domain, and macro-averaged to obtain domain-level results.


\section{Results}

This section presents a comprehensive analysis of QE performance across domains, languages, prompting strategies, and parameter-efficient adaptation methods. We begin by analysing the adapter-based approaches using \alope, which highlights how Transformer layer selection affects QE performance. We then contextualise these findings through domain-wise comparisons against prompt-only baselines, enabling a detailed understanding of when and why lightweight adaptation becomes necessary.

\begin{table*}[t]
\centering
\normalsize
\setlength{\tabcolsep}{1.3pt}
\renewcommand{\arraystretch}{1.5}

\resizebox{\textwidth}{!}{
\begin{NiceTabular}{l|l|cccccc|cccccc|cccccc}
\toprule
\textbf{} & \textbf{Layer} &
\multicolumn{6}{c|}{\textbf{\boldmath$R=32$, $\alpha=16$}} &
\multicolumn{6}{c|}{\textbf{\boldmath$R=64$, $\alpha=32$}} &
\multicolumn{6}{c}{\textbf{\boldmath$R=128$, $\alpha=32$}} \\
\cmidrule(lr){3-8}\cmidrule(lr){9-14}\cmidrule(lr){15-20}
& &
\textbf{En-Hi} & \textbf{En-Mr} & \textbf{En-Ta} & \textbf{En-Te} & \textbf{En-Gu} & \textbf{Avg} &
\textbf{En-Hi} & \textbf{En-Mr} & \textbf{En-Ta} & \textbf{En-Te} & \textbf{En-Gu} & \textbf{Avg} &
\textbf{En-Hi} & \textbf{En-Mr} & \textbf{En-Ta} & \textbf{En-Te} & \textbf{En-Gu} & \textbf{Avg} \\
\midrule

\Block{4-1}{%
  \parbox[c][3.0cm][c]{0.5cm}{%
    \centering
    \rotatebox[origin=c]{90}{\textbf{\general}}%
  }%
}
& L $-1$  & 0.363 & -0.026 & 0.385 & 0.180 & 0.190 & 0.218 &\textbf{ 0.476} & 0.155 & 0.605 & 0.267 & 0.435 & $\dagger$0.388 & 0.384 & 0.105 & 0.524 & 0.196 & 0.391 & 0.320 \\
& L $-7$  & 0.171 & \textbf{0.157}  & 0.389 & 0.176 & 0.212 & 0.221 & 0.213 & 0.059 & 0.469 & 0.252 & 0.371 & 0.273 & 0.057 & 0.057 & 0.346 & 0.134 & 0.320 & 0.183 \\
& L $-9$  & 0.123 & 0.090  & 0.531 & 0.163 & 0.458 & 0.273 & 0.311 & 0.038 & 0.531 & 0.279 &\textbf{ 0.485} & 0.329 & 0.051 & 0.107 & 0.300 & 0.061 & 0.343 & 0.172 \\
& L $-11$ & 0.178 & 0.061  & 0.492 & -0.029 & 0.484 & 0.237 & 0.321 & 0.034 & 0.568 & -0.021 & 0.480 & 0.276 & 0.308 & 0.007 & \textbf{0.610} & \textbf{0.292} & 0.457 & 0.335 \\

\midrule
\Block{4-1}{%
  \parbox[c][3.0cm][c]{0.5cm}{%
    \centering
    \rotatebox[origin=c]{90}{\textbf{\healthcare}}%
  }%
}
& L $-1$  & \textbf{0.520} & 0.025 & 0.201 & NA & 0.377 & 0.281 & 0.382 & 0.033 & 0.248 & NA & 0.490 & 0.288 & 0.346 & 0.156 & 0.165 & NA & 0.452 & 0.280 \\
& L $-7$  & 0.168 & 0.150 & 0.154 & NA & 0.346 & 0.204 & 0.314 & 0.021 & 0.363 & NA & \textbf{0.532} & 0.307 & 0.283 & 0.182 & 0.392 & NA & 0.477 & $\dagger$0.333 \\
& L $-9$  & 0.110 & 0.096 & 0.086 & NA & 0.255 & 0.137 & 0.262 & 0.132 & 0.414 & NA & 0.417 & 0.306 & 0.161 & 0.019 & 0.321 & NA & 0.358 & 0.215 \\
& L $-11$ & 0.123 & 0.050 & 0.352 & NA & 0.435 & 0.240 & 0.265 & \textbf{0.192} & 0.320 & NA & 0.469 & 0.311 & 0.346 & -0.042 & \textbf{0.415} & NA & 0.511 & 0.308 \\

\midrule
\Block{4-1}{%
  \parbox[c][3.0cm][c]{0.5cm}{%
    \centering
    \rotatebox[origin=c]{90}{\textbf{\legal}}%
  }%
}
& L $-1$  & NA & NA & -0.091 & -0.009 & 0.002 & -0.033 & NA & NA & 0.518 & 0.071 & 0.259 & 0.283 & NA & NA & 0.380 & 0.136 & 0.355 & 0.290 \\
& L $-7$  & NA & NA & 0.379 & 0.052 & 0.367 & 0.266 & NA & NA & 0.443 & 0.059 & 0.366 & 0.289 & NA & NA & 0.478 & 0.124 & 0.280 & 0.294 \\
& L $-9$  & NA & NA & 0.558 & 0.143 & 0.305 & 0.335 & NA & NA & 0.521 & 0.079 & 0.278 & 0.293 & NA & NA & 0.432 & 0.107 & 0.283 & 0.274 \\
& L $-11$ & NA & NA & 0.489 & 0.043 & 0.243 & 0.258 & NA & NA & 0.580 & \textbf{0.267} & \textbf{0.445} & 0.430 & NA & NA & \textbf{0.581} & 0.267 & 0.445 & $\dagger$0.431 \\

\midrule
\Block{4-1}{%
  \parbox[c][3.0cm][c]{0.5cm}{%
    \centering
    \rotatebox[origin=c]{90}{\textbf{\tourism}}%
  }%
}
& L $-1$  & -0.153 & -0.009 & NA & 0.002 & NA & -0.053 & -0.080 & 0.029 & NA & 0.065 & NA & 0.047 & -0.061 & 0.045 & NA & 0.080 & NA & 0.063 \\
& L $-7$  & -0.104 & 0.516 & NA & 0.019 & NA & 0.144 & 0.125 & 0.566 & NA & 0.154 & NA & 0.282 & 0.180 & 0.600 & NA & 0.210 & NA & 0.330 \\
& L $-9$  & -0.082 & 0.579 & NA & -0.050 & NA & 0.149 & 0.288 & 0.596 & NA & 0.189 & NA & 0.357 & 0.330 & 0.640 & NA & \textbf{0.220} & NA & 0.397 \\
& L $-11$ & 0.298 & 0.633 & NA & 0.183 & NA & 0.371 & 0.298 & 0.633 & NA & 0.183 & NA & 0.371 & \textbf{0.350} &\textbf{ 0.670} & NA & 0.205 & NA & $\dagger$0.408 \\

\bottomrule
\end{NiceTabular}
}
\caption{Spearman’s ($\rho$) scores obtained for \alope layer-wise (L) experiments with LoRA across different domains and language pairs. The bolded values represent the highest spearman scores obtained for each language pair in each domain. The ($\dagger$) represents the highest average obtained in each domain. `NA' indicates that data from that specific domain and language pair is unavailable.} 
\label{tab:alope_lora_layerwise_full}
\end{table*}

\subsection{Layer-wise Analysis with ALOPE}
Table~\ref{tab:alope_lora_layerwise_full} shows that, for the \general domain, the highest Spearman correlations for most language pairs are achieved at intermediate Transformer layers when using \alope with LoRA. The same pattern is observed consistently across \healthcare, \legal, and \tourism domains. This behaviour is mirrored by \alope with LoRMA (Table~\ref{tab:lorma_layerwise_full}), where peak correlations for individual language pairs are likewise concentrated at intermediate layers, particularly Layers $-7$, $-9$, and $-11$.

To summarise the overall trends, we visualise the domain-level average Spearman correlations from Tables~\ref{tab:alope_lora_layerwise_full} and~\ref{tab:lorma_layerwise_full} in Figures~\ref{fig:ALOPE} and~\ref{fig:LORMA} in Appendix~\ref{app:average_ALOPE_figures}. Each figure plots the average correlation computed across the five language pairs for a given domain—against the Transformer layer from which the regression head extracts representations. Across all domains, the averaged results reveal a clear and consistent pattern: intermediate Transformer layers, particularly Layers $-9$ and $-11$, yield substantially higher Spearman correlations than the final layer (Layer $-1$). This trend is stable across domains and adapter variants, and aligns with prior findings of \alope~\cite{sindhujan-etal-2025-alope}. These results support the hypothesis that QE-relevant signals are more robustly encoded in intermediate representations for English$\rightarrow$Indic language pairs, while final-layer representations are more specialised for next-token prediction and instruction-following objectives.

Further analysis of the averaged correlations indicates that \alope with LoRMA introduces a stabilising effect across layers (Appendix~\ref{app:average_ALOPE_figures}, Figure~\ref{fig:LORMA}). Compared to LoRA, LoRMA produces smoother layer-wise behaviour, mitigating extremely low correlations at shallow layers (e.g., Layers $-1$ and $-7$) and reducing variance between adjacent layers across most configurations. This stabilisation is most pronounced in the \general and \legal domains, where performance becomes less sensitive to the exact choice of layer.

Despite these stability gains, LoRMA obtains competitive performance to LoRA in  the \tourism (0.408 vs.\ 0.404) domain. In \healthcare and \legal domains, LoRA consistently delivers higher correlations. These findings highlight a principled trade-off: LoRA is preferable when maximising ranking accuracy is the primary objective, whereas LoRMA offers increased robustness when deployment constraints limit precise layer selection.

\paragraph{Adapter configurations.} \alope has different adapter configurations as explained in the section ~\ref{subsec:ALOPE}. Lower-rank adapters with $R=32$ consistently underfit across several domains, achieving substantially lower correlations than higher-rank configurations. Increasing adapter capacity to $R=128$ occasionally improves the highest correlations but introduces instability and higher variance, particularly for shallow layers. In contrast, $R=64$ with $\alpha=32$ consistently provides the best balance between expressive capacity and robustness across domains. 

\begin{table*}[ht]
\centering
\normalsize
\setlength{\tabcolsep}{3pt}
\renewcommand{\arraystretch}{1.35}
\begin{adjustbox}{max width=\textwidth}
\begin{NiceTabular}{l|l|cccccc|cccccc}
\toprule
\textbf{} & \textbf{Layer} &
\multicolumn{6}{c|}{\textbf{\boldmath$R=64$, $\alpha=32$}} &
\multicolumn{6}{c}{\textbf{\boldmath$R=128$, $\alpha=32$}} \\
\cmidrule(lr){3-8}\cmidrule(lr){9-14}
& &
\textbf{En-Hi} & \textbf{En-Mr} & \textbf{En-Ta} & \textbf{En-Te} & \textbf{En-Gu} & \textbf{Avg} &
\textbf{En-Hi} & \textbf{En-Mr} & \textbf{En-Ta} & \textbf{En-Te} & \textbf{En-Gu} & \textbf{Avg} \\
\midrule

\Block{4-1}{%
  \parbox[c][3.0cm][c]{0.5cm}{%
    \centering
    \rotatebox[origin=c]{90}{\textbf{\general}}%
  }%
}
& L $-1$  & 0.299 & 0.068 & 0.481 & 0.114 & 0.298 & 0.252 & 0.276 & 0.092 & 0.330 & 0.059 & 0.295 & 0.210 \\
& L $-7$  & 0.324 &\textbf{ 0.314} & 0.438 & 0.087 & 0.381 & 0.309 & \textbf{0.392} & 0.081 & \textbf{0.507} & 0.155 & \textbf{0.421} & 0.311 \\
& L $-9$  & 0.255 & 0.265 & 0.473 & 0.070 & 0.362 & 0.285 & 0.278 & 0.283 & 0.484 & 0.101 & 0.386 & 0.306 \\
& L $-11$ & 0.337 & 0.267 & 0.483 & \textbf{0.168} & 0.391 & $\dagger$0.329 & 0.343 & 0.259 & 0.409 & 0.086 & 0.404 & 0.300 \\
\midrule

\Block{4-1}{%
  \parbox[c][3.0cm][c]{0.5cm}{%
    \centering
    \rotatebox[origin=c]{90}{\textbf{\healthcare}}%
  }%
}
& L $-1$  & -0.316 & -0.021 & -0.381 & NA & 0.420 & -0.075 & -0.076 & -0.025 & 0.374 & NA & \textbf{0.512} & 0.196 \\
& L $-7$  & 0.262 & 0.124 & 0.405 & NA & 0.495 & 0.322 & \textbf{0.365} & -0.017 & \textbf{0.413} & NA & 0.457 & 0.305 \\
& L $-9$  & 0.077 & 0.074 & 0.400 & NA & 0.453 & 0.251 & 0.343 & \textbf{0.131} & 0.277 & NA & 0.490 & 0.310 \\
& L $-11$ & 0.315 & 0.016 & 0.412 & NA & 0.482 & 0.306 & 0.351 & 0.105 & 0.392 & NA & 0.467 & $\dagger$0.329 \\
\midrule

\Block{4-1}{%
  \parbox[c][3.0cm][c]{0.5cm}{%
    \centering
    \rotatebox[origin=c]{90}{\textbf{\legal}}%
  }%
}
& L $-1$  & NA & NA & 0.509 & 0.073 & 0.182 & 0.255 & NA & NA & \textbf{0.515} & 0.104 & 0.173 & 0.264 \\
& L $-7$  & NA & NA & 0.488 & 0.092 & 0.193 & 0.258 & NA & NA & 0.478 & 0.077 & 0.214 & 0.256 \\
& L $-9$  & NA & NA & 0.488 & 0.112 & \textbf{0.240} & $\dagger$0.280 & NA & NA & 0.430 & \textbf{0.125} & 0.216 & 0.257 \\
& L $-11$ & NA & NA & 0.424 & 0.058 & 0.178 & 0.220 & NA & NA & 0.480 & 0.111 & 0.110 & 0.234 \\
\midrule

\Block{4-1}{%
  \parbox[c][3.0cm][c]{0.5cm}{%
    \centering
    \rotatebox[origin=c]{90}{\textbf{\tourism}}%
  }%
}
& L $-1$  & -0.315 & 0.462 & NA & -0.082 & NA & 0.022 & -0.354 & -0.013 & NA & -0.171 & NA & -0.179 \\
& L $-7$  & 0.436 & 0.493 & NA & 0.184 & NA & 0.371 & 0.446 & 0.502 & NA & \textbf{0.197} & NA & 0.382 \\
& L $-9$  & 0.453 & \textbf{0.532} & NA & 0.227 & NA & $\dagger$0.404 & \textbf{0.465} & 0.445 & NA & 0.142 & NA & 0.351 \\
& L $-11$ & 0.389 & 0.512 & NA & 0.074 & NA & 0.325 & 0.423 & 0.468 & NA & 0.164 & NA & 0.352 \\

\bottomrule
\end{NiceTabular}
\end{adjustbox}

\caption{Spearman’s ($\rho$) scores obtained for \alope layer-wise (L) experiments with LoRMA across different domains and language pairs. The bolded values represent the highest spearman scores obtained for each language pair in each domain. The ($\dagger$) represents the highest average obtained in each domain. `NA' indicates that a language pair is unavailable for that specific domain.}
\label{tab:lorma_layerwise_full}
\end{table*}

\subsection{Prompt-only QE Baselines}

\begin{table*}[t]
\centering
\small
\setlength{\tabcolsep}{4pt}
\resizebox{\textwidth}{!}{
\begin{tabular}{lllcccccc}
\toprule
\textbf{Domain} & \textbf{Prompt setting} & \textbf{Model family} &
\textbf{En-Hi} & \textbf{En-Mr} & \textbf{En-Ta} & \textbf{En-Te} & \textbf{En-Gu} & \textbf{Average} \\
\midrule

\multirow{7}{*}{\textbf{\general}}
& \multirow{2}{*}{Zero-shot} & Closed & 0.424 & \textbf{0.597} & 0.848 & 0.392 & \textbf{0.924} & $\dagger$0.637 \\
&                            & Open   & 0.390 & -0.058 & 0.772 & 0.382 & 0.812 & 0.460 \\
& \multirow{2}{*}{Few-shot + Guidelines} & Closed & 0.475 & 0.582 & 0.886 & 0.238 & 0.776 & 0.591 \\
&                                       & Open   & 0.408 & 0.038 & 0.832 & 0.442 & 0.849 & 0.514 \\
& \multirow{2}{*}{Few-shot (No Guidelines)} & Closed & \textbf{0.563} & 0.314 & \textbf{0.940} & -0.031 & 0.860 & 0.529 \\
&                                           & Open   & 0.418 & 0.375 & 0.867 & \textbf{0.486} & 0.752 & 0.580 \\
\midrule

\multirow{7}{*}{\textbf{\healthcare}}
& \multirow{2}{*}{Zero-shot} & Closed & 0.126 & 0.814 & 0.366 & NA & 0.346 & 0.413 \\
&                            & Open   & 0.569 & 0.389 & \textbf{0.603} & NA & \textbf{0.494} & 0.514 \\
& \multirow{2}{*}{Few-shot + Guidelines} & Closed & 0.415 & 0.669 & -0.040 & NA & 0.100 & 0.286 \\
&                                       & Open   & 0.447 & \textbf{0.884} & 0.411 & NA & 0.211 & 0.488 \\
& \multirow{2}{*}{Few-shot (No Guidelines)} & Closed & 0.585 & 0.786 & 0.168 & NA & 0.458 & 0.385 \\
&                                           & Open   & \textbf{0.611} & \textbf{0.884} & 0.422 & NA & 0.398 & $\dagger$0.579 \\
\midrule

\multirow{7}{*}{\textbf{\legal}}
& \multirow{2}{*}{Zero-shot} & Closed & NA & NA & \textbf{0.749} & 0.109 & 0.677 & 0.512 \\
&                            & Open   & NA & NA & 0.418 & 0.287 & 0.265 & 0.323 \\
& \multirow{2}{*}{Few-shot + Guidelines} & Closed & NA & NA & 0.717 & 0.530 & 0.699 & $\dagger$0.649 \\
&                                       & Open   & NA & NA & 0.418 & 0.528 & \textbf{0.727} & 0.558 \\
& \multirow{2}{*}{Few-shot (No Guidelines)} & Closed & NA & NA & 0.475 & 0.230 & 0.475 & 0.393 \\
&                                           & Open   & NA & NA & 0.737 & \textbf{0.687} & 0.473 & 0.632 \\
\midrule

\multirow{7}{*}{\textbf{\tourism}}
& \multirow{2}{*}{Zero-shot} & Closed & 0.416 & 0.474 & NA & 0.217 & NA & 0.369 \\
&                            & Open   & 0.613 & 0.689 & NA & \textbf{0.636} & NA & $\dagger$0.646 \\
& \multirow{2}{*}{Few-shot + Guidelines} & Closed & 0.502 & 0.679 & NA & 0.472 & NA & 0.551 \\
&                                       & Open   & 0.158 & \textbf{0.702} & NA & 0.583 & NA & 0.481 \\
& \multirow{2}{*}{Few-shot (No Guidelines)} & Closed & 0.509 & 0.685 & NA & 0.397 & NA & 0.530 \\
&                                           & Open   & \textbf{0.737} & 0.687 & NA & 0.473 & NA & 0.632 \\
\bottomrule
\end{tabular}
}
\caption{Reports best Spearman’s correlation ($\rho$) achieved for each domain and language pair across all evaluated models, including both open- and closed-weight LLMs. The bolded values represent the highest spearman scores obtained for each language pair in each domain. The ($\dagger$) represents the highest average obtained in each domain. Detailed model-wise results are provided in Appendices~\ref{app:results_zero_prompting},~\ref{app:results_few_guideline_prompting}, and~\ref{app:results_few_without_guideline_prompting}. }
\label{tab:prompt_qe_results}
\end{table*}

Table ~\ref{tab:prompt_qe_results} shows the best Spearman scores obtained with prompt-only baselines for each language pair and domain with closed- and open-weight models (Detailed model-wise results are reported in Appendix~\ref{app:results_zero_prompting}, ~\ref{app:results_few_guideline_prompting},~\ref{app:results_few_without_guideline_prompting}). 

Closed-weight models consistently provide strong performance across domains, even under zero-shot prompting. This behaviour, which yields competitive scores with minimal prompt engineering, can be attributed to the large scale of the Gemini models and their extensive pre-training. Open-weight models exhibit substantially weaker and more variable behaviour under prompt-only evaluation, particularly in \healthcare and \legal. Overall, considering both open- and closed-weights models, few-shot prompting improves performance, and guideline-based prompts further stabilise the Spearman scores.

\subsection{ALOPE vs.\ Prompt-only Approaches}

Since \alope was evaluated exclusively with LLaMA-3.2-3B, our comparative analysis focuses on prompt-only results obtained with the same backbone model (Appendix~\ref{app:domain_prompt_vs_alope}). Across the majority of language pairs and domains, \alope consistently achieves higher Spearman correlations than prompt-only prompting strategies. These results demonstrate that \alope provides a practical and effective approach for improving QE, even when applied to smaller open-weight LLMs with quantization and lightweight adapters, which substantially reduce parameter count, model size, and computational cost compared to large closed-weight models. While \alope is applicable to other open-weight architectures, extending this analysis to additional models is left for future work.

\subsection{Domain-specific Observations}

This section analyses performance differences between the best prompt-only baselines (Table~\ref{tab:prompt_qe_results}) and the \alope-based approaches (Tables~\ref{tab:alope_lora_layerwise_full} and~\ref{tab:lorma_layerwise_full}). Across all settings, the \general domain consistently achieves the highest correlations, reflecting its broader linguistic coverage and lower terminological complexity compared to specialised domains.

\healthcare presents a mixed picture for \alope's effectiveness. While prompt-only baselines achieve strong correlations for several language pairs, \alope with intermediate-layer adaptation shows limited improvements and occasionally underperforms these baselines. Notable exceptions include English$\rightarrow$Gujarati, where \alope achieves competitive performance. This suggests that \healthcare QE benefits more from strong prompting strategies with closed-weight models than from lightweight adapter-based methods for most language pairs. When \alope does improve performance, gains are strongest at the Transformer Layers $-7$ to $-11$ (Table~\ref{tab:alope_lora_layerwise_full}) for this domain.

\legal remains the most challenging domain overall. \alope shows selective improvements, particularly for English$\rightarrow$Tamil ($\rho$ = 0.581, outperforming open-weight prompt baselines), though absolute correlations remain lower than other domains. Performance gains are configuration-sensitive, reflecting the strict semantic requirements of \legal text.

\tourism exhibits an irregular pattern where zero-shot prompting with open-weight models achieves surprisingly strong performance (average $\rho$ = 0.646), often matching or exceeding average Spearman's obtained with stronger closed-weight models and \alope. For English$\rightarrow$Marathi, \alope achieves $\rho$ = 0.670, which is competitive but does not surpass the best closed-weight prompt-only baseline (few-shot with guideline: $\rho$ = 0.702). Gains for other language pairs level off quickly, consistent with the entity-heavy and descriptive nature of \tourism content, where surface-level fluency and entity preservation may be adequately captured by prompting alone.

\section{Discussion}

Closed-weight models achieve strong and stable correlations even under zero-shot prompting, reflecting their extensive instruction tuning and multilingual pre-training. Guideline-anchored prompts further improve robustness by clarifying scale semantics and reducing mid-range score compression \citep{mishra2022cross,zhao2021calibrate}. As a result, closed-weight models represent the most reliable option when API access is available. In contrast, open-weight models exhibit substantially weaker and more variable performance under prompt-only evaluation, with zero-shot prompting often yielding near-zero or negative correlations. This highlights the inherent limitations of prompt engineering alone for smaller models, particularly in high-risk domains \citep{info16100916}.

\alope provides a targeted remedy for this gap, but its effectiveness is strongly domain-dependent. In the \legal domain, \alope substantially improves performance, underscoring the importance of domain-specific adaptation for semantically precise content. In contrast, gains in \healthcare are limited, suggesting that this domain benefits more from broad pre-training coverage of medical terminology in large closed-weight models than from lightweight adapter-based fine-tuning. \general and \tourism domains exhibit intermediate behaviour, indicating that the utility of \alope depends on the interaction between domain complexity and pre-training corpus characteristics. Incorporating LoRMA within \alope further introduces stability-oriented regularisation, yielding smoother layer-wise behaviour and mitigating lower correlations at shallow layers \citep{sindhujan-etal-2025-alope}.

The layer-wise superiority of intermediate representations (Layers -9, -11) holds consistently across all five languages, suggesting this reflects fundamental properties of how multilingual LLMs encode cross-lingual semantic alignment rather than language-specific artifacts \citep{kargaran2025layerwise}. This consistency across typologically diverse languages strengthens the generalisability of intermediate-layer adaptation for QE in low-resource settings.

Taken together, our results motivate a conditional deployment strategy. When access to closed-weight models is feasible, guideline-anchored prompting offers the most reliable solution. When deployment constraints such as cost, latency, or privacy preclude such access, \alope with LoRA provides a lightweight and effective alternative, particularly for semantically complex domains such as \legal. This underscores the importance of empirical validation prior to adopting adaptation strategies, as their effectiveness depends on domain-specific interactions with pre-training data.

\section{Conclusion}
This work investigates domain-specific quality estimation for English$\rightarrow$Indic translation across \healthcare, \legal, \tourism, and \general domains, covering five language pairs (Hindi, Marathi, Tamil, Telugu, Gujarati). We systematically compare prompt-only approaches with parameter-efficient \alope adaptation, revealing that closed-weight models with guideline-anchored prompting achieve robust performance without parameter updates, while open-weight models exhibit substantial fragility under prompt-only evaluation. \alope demonstrates reasonable QE performance even with smaller open-source LLMs when intermediate Transformer layer embeddings are utilised for quality estimation. Our findings support a conditional deployment strategy: prioritise closed-weight prompting when API access is viable; apply \alope with LoRA for open-weight models in resource-constrained environment; and use \alope with LoRMA when precise layer tuning is constrained. Future work should investigate multi-layer fusion approaches and interpretability techniques to understand which linguistic phenomena drive layer-specific improvements.

\section{Limitations}
Our study is limited to English$\rightarrow$Indic language pairs across four domains, due to the limited availability of QE data for other language pairs and domains. Further work is needed to examine the generalisability of these findings to other language families and more specialised technical domains. In addition, all \alope fine-tuning experiments use a relatively small backbone model (LLaMA-3.2-3B Instruct) due to computational constraints, and results may differ when scaling to larger open-weight models.










\bibliography{custom}
\clearpage

\appendix
\onecolumn

\section{Prompt Templates for Prompt-only Approaches} \label{app:prompts}

 \begin{figure*}[ht!]
 \centering
 \includegraphics[width=0.9\textwidth, keepaspectratio]{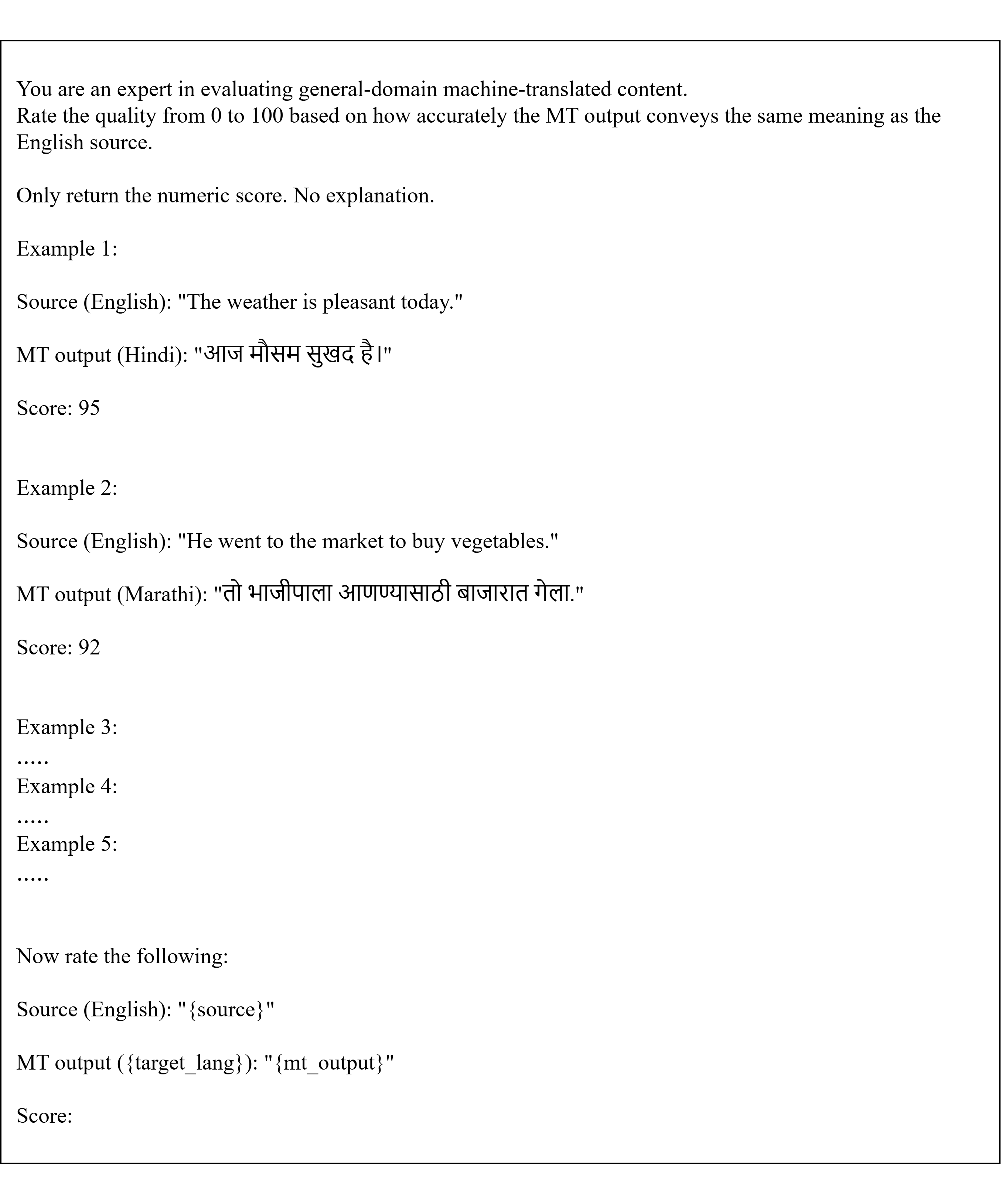}
\caption{ Few-shot QE Prompt (Without Guidelines)}
\label{fig:fewshot_fewshot_without_guideline} 
\end{figure*}
\clearpage 

 \begin{figure*}[htbp]
 \centering
 \includegraphics[width=0.9\textwidth, keepaspectratio]{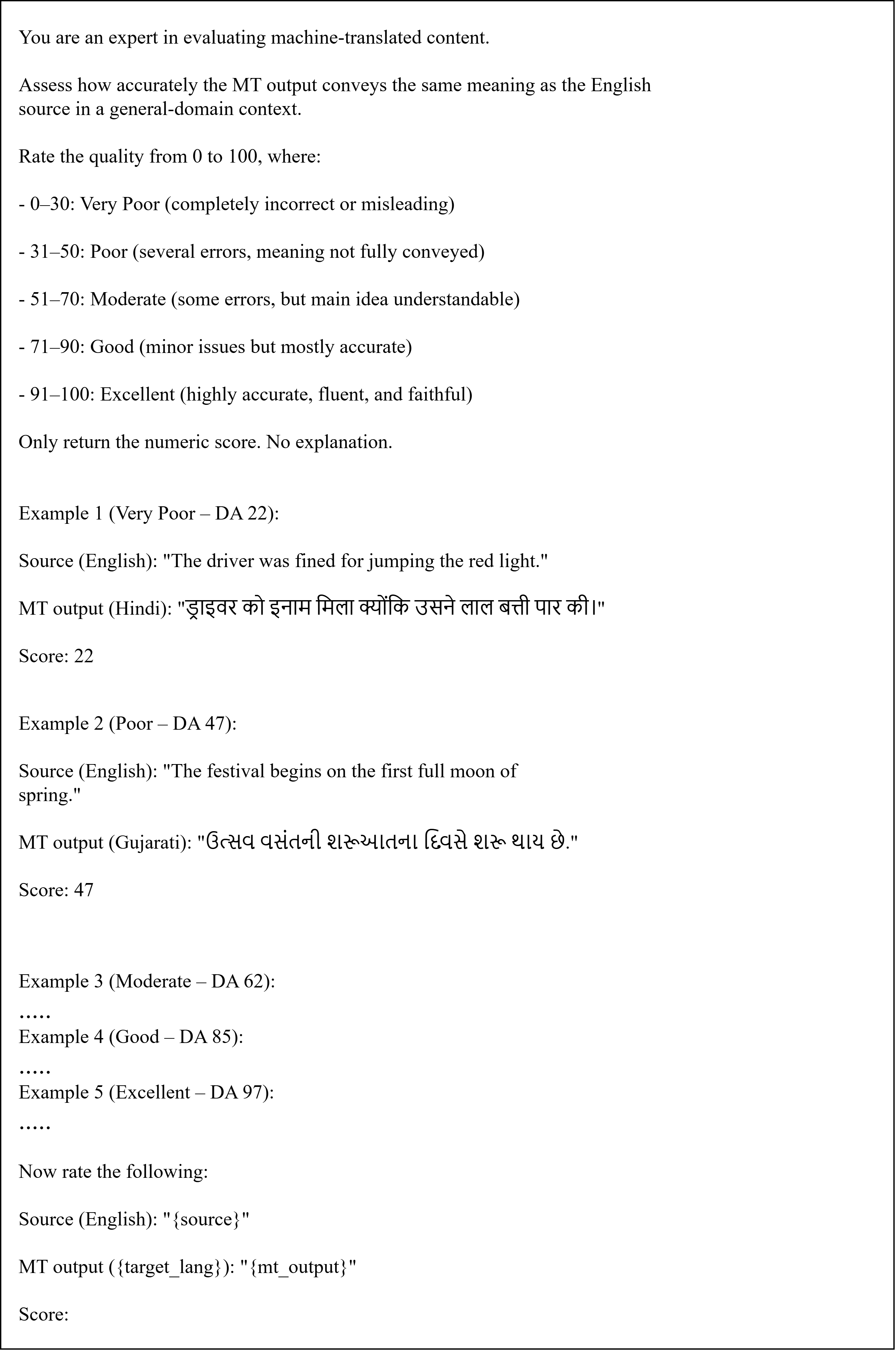}
\caption{ Few-shot QE Prompt (With Guidelines)}
\label{fig:fewshot_with_guideline} 
\end{figure*}
\clearpage 

 \begin{figure*}[ht]
 \centering
 \includegraphics[width=0.8\textwidth, keepaspectratio]{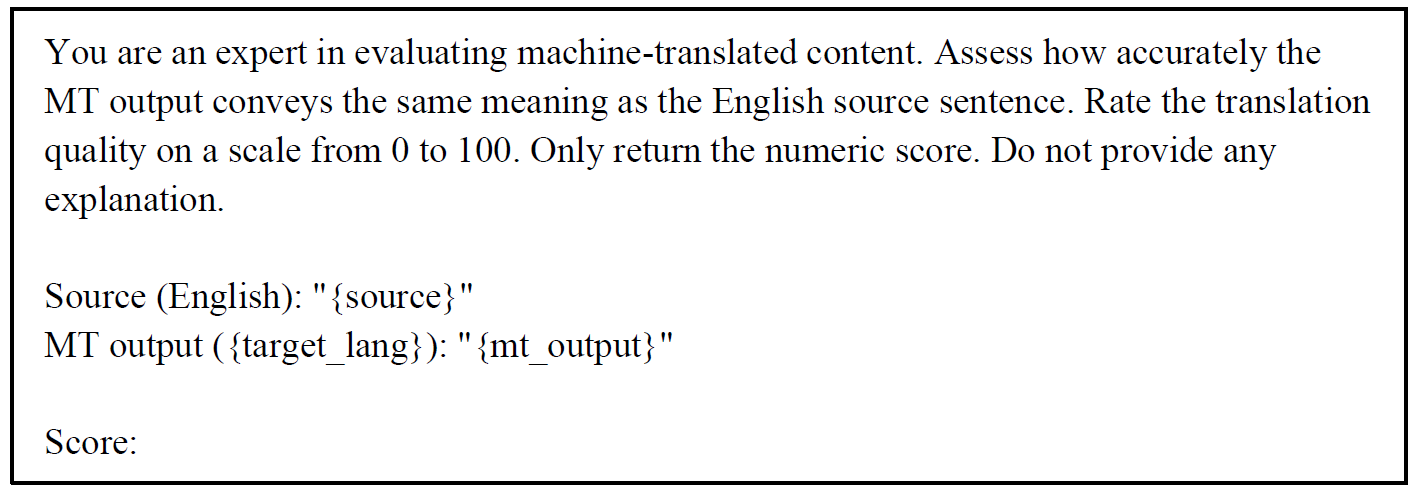}
\caption{ Zero-shot prompt}
\label{fig:zero-shot} 
\end{figure*}

\section{Prompt Templates for ALOPE} \label{app:results_zero_prompting}

\begin{figure*}[ht]
 \centering
 \includegraphics[width=0.8\textwidth, keepaspectratio]{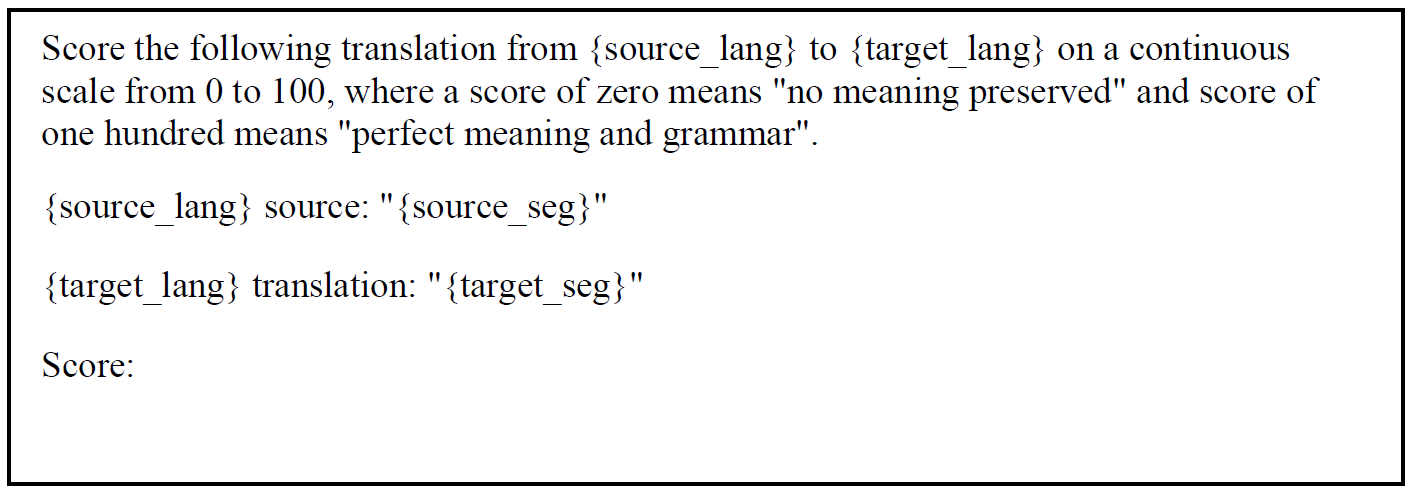}
\caption{ ALOPE prompt}
\label{fig:ALOPE_prompt} 
\end{figure*}
\clearpage

\section{Pearson's Correlation Scores obtained for ALOPE with LoRA } \label{app:table-ALOPE-Lora}

\begin{table*}[ht]
\centering
\normalsize
\setlength{\tabcolsep}{3pt}
\renewcommand{\arraystretch}{1.35}
\begin{adjustbox}{max width=\textwidth}
\begin{NiceTabular}{l|l|cccccc|cccccc|cccccc}
\toprule
\textbf{} & \textbf{Layer} &
\multicolumn{6}{c|}{\textbf{\boldmath$R=32$, $\alpha=16$}} &
\multicolumn{6}{c|}{\textbf{\boldmath$R=64$, $\alpha=32$}} &
\multicolumn{6}{c}{\textbf{\boldmath$R=128$, $\alpha=32$}} \\
\cmidrule(lr){3-8}\cmidrule(lr){9-14}\cmidrule(lr){15-20}
& &
\textbf{En-Hi} & \textbf{En-Mr} & \textbf{En-Ta} & \textbf{En-Te} & \textbf{En-Gu} & \textbf{Avg} &
\textbf{En-Hi} & \textbf{En-Mr} & \textbf{En-Ta} & \textbf{En-Te} & \textbf{En-Gu} & \textbf{Avg} &
\textbf{En-Hi} & \textbf{En-Mr} & \textbf{En-Ta} & \textbf{En-Te} & \textbf{En-Gu} & \textbf{Avg} \\
\midrule

\Block{4-1}{%
  \parbox[c][3.0cm][c]{0.5cm}{\centering\rotatebox[origin=c]{90}{\textbf{\general}}}
}
& L $-1$  & 0.400 & -0.025 & 0.305 & 0.092 & 0.221 & 0.199
          & 0.628 & 0.154 & 0.680 & 0.229 & 0.538 & 0.446
          & 0.597 & 0.111 & 0.679 & 0.181 & 0.594 & 0.433 \\
& L $-7$  & 0.538 & 0.117 & 0.645 & 0.244 & 0.272 & 0.363
          & 0.488 & 0.053 & 0.674 & 0.216 & 0.562 & 0.399
          & 0.306 & 0.043 & 0.527 & 0.177 & 0.471 & 0.305 \\
& L $-9$  & 0.492 & 0.081 & 0.677 & 0.250 & 0.679 & 0.436
          & 0.621 & 0.077 & 0.716 & 0.305 & 0.663 & 0.476
          & 0.328 & 0.095 & 0.549 & 0.131 & 0.511 & 0.323 \\
& L $-11$ & 0.539 & 0.080 & 0.720 & 0.016 & 0.687 & 0.408
          & 0.582 & 0.019 & 0.674 & 0.031 & 0.637 & 0.389
          & 0.585 & 0.039 & 0.647 & 0.257 & 0.583 & 0.422 \\
\midrule

\Block{4-1}{%
  \parbox[c][3.0cm][c]{0.5cm}{\centering\rotatebox[origin=c]{90}{\textbf{\healthcare}}}
}
& L $-1$  & 0.589 & 0.001 & 0.405 & NA & 0.354 & 0.337
          & 0.635 & 0.051 & 0.494 & NA & 0.440 & 0.405
          & 0.707 & 0.164 & 0.464 & NA & 0.426 & 0.440 \\
& L $-7$  & 0.514 & 0.158 & 0.502 & NA & 0.381 & 0.389
          & 0.682 & 0.008 & 0.523 & NA & 0.499 & 0.428
          & 0.708 & 0.180 & 0.617 & NA & 0.465 & 0.493 \\
& L $-9$  & 0.554 & 0.076 & 0.435 & NA & 0.342 & 0.352
          & 0.688 & 0.094 & 0.629 & NA & 0.405 & 0.454
          & 0.616 & 0.002 & 0.608 & NA & 0.358 & 0.396 \\
& L $-11$ & 0.625 & 0.015 & 0.611 & NA & 0.425 & 0.419
          & 0.614 & 0.193 & 0.630 & NA & 0.469 & 0.477
          & 0.685 & -0.018 & 0.602 & NA & 0.496 & 0.441 \\
\midrule

\Block{4-1}{%
  \parbox[c][3.0cm][c]{0.5cm}{\centering\rotatebox[origin=c]{90}{\textbf{\legal}}}
}
& L $-1$  & NA & NA & 0.059 & 0.002 & 0.011 & 0.024
          & NA & NA & 0.468 & 0.040 & 0.318 & 0.276
          & NA & NA & 0.430 & 0.100 & 0.382 & 0.304 \\
& L $-7$  & NA & NA & 0.502 & 0.077 & 0.457 & 0.345
          & NA & NA & 0.496 & 0.088 & 0.421 & 0.335
          & NA & NA & 0.460 & 0.093 & 0.355 & 0.303 \\
& L $-9$  & NA & NA & 0.551 & 0.142 & 0.387 & 0.360
          & NA & NA & 0.459 & 0.059 & 0.334 & 0.284
          & NA & NA & 0.452 & 0.081 & 0.354 & 0.295 \\
& L $-11$ & NA & NA & 0.447 & 0.001 & 0.334 & 0.261
          & NA & NA & 0.621 & 0.215 & 0.536 & 0.457
          & NA & NA & 0.621 & 0.215 & 0.536 & 0.457 \\
\midrule

\Block{4-1}{%
  \parbox[c][3.0cm][c]{0.5cm}{\centering\rotatebox[origin=c]{90}{\textbf{\tourism}}}
}
& L $-1$  & 0.059 & 0.002 & NA & 0.011 & NA & 0.024
          & 0.119 & 0.064 & NA & 0.083 & NA & 0.089
          & 0.130 & 0.075 & NA & 0.095 & NA & 0.100 \\
& L $-7$  & 0.061 & 0.467 & NA & 0.145 & NA & 0.224
          & 0.220 & 0.544 & NA & 0.286 & NA & 0.350
          & 0.260 & 0.580 & NA & 0.320 & NA & 0.387 \\
& L $-9$  & 0.103 & 0.557 & NA & 0.069 & NA & 0.243
          & 0.317 & 0.614 & NA & 0.235 & NA & 0.389
          & 0.360 & 0.660 & NA & 0.270 & NA & 0.430 \\
& L $-11$ & 0.485 & 0.641 & NA & 0.219 & NA & 0.448
          & 0.485 & 0.641 & NA & 0.219 & NA & 0.448
          & 0.525 & 0.685 & NA & 0.245 & NA & 0.485 \\

\bottomrule
\end{NiceTabular}
\end{adjustbox}

\caption{ALOPE (LoRA) layer-wise (L) Pearson’s ($r$) scores obtained across domains and language pairs for different adapter configurations.}
\label{tab:alope_layer_analysis}
\end{table*}

\clearpage
\section{Pearson's Correlation Scores obtained for ALOPE with LoRMA} \label{app:table-ALOPE-LoRMA}

\begin{table*}[ht]
\centering
\normalsize
\setlength{\tabcolsep}{3pt}
\renewcommand{\arraystretch}{1.35}
\begin{adjustbox}{max width=\textwidth}
\begin{NiceTabular}{l|l|cccccc|cccccc}
\toprule
\textbf{} & \textbf{Layer} &
\multicolumn{6}{c|}{\textbf{\boldmath$R=64$, $\alpha=32$}} &
\multicolumn{6}{c}{\textbf{\boldmath$R=128$, $\alpha=32$}} \\
\cmidrule(lr){3-8}\cmidrule(lr){9-14}
& &
\textbf{En-Hi} & \textbf{En-Mr} & \textbf{En-Ta} & \textbf{En-Te} & \textbf{En-Gu} & \textbf{Avg} &
\textbf{En-Hi} & \textbf{En-Mr} & \textbf{En-Ta} & \textbf{En-Te} & \textbf{En-Gu} & \textbf{Avg} \\
\midrule

\Block{4-1}{%
  \parbox[c][3.0cm][c]{0.5cm}{\centering\rotatebox[origin=c]{90}{\textbf{\general}}}
}
& L $-1$  & 0.262 & 0.030 & 0.552 & 0.053 & 0.430 & 0.259
          & 0.217 & 0.120 & 0.383 & -0.007 & 0.311 & 0.205 \\
& L $-7$  & 0.382 & 0.315 & 0.557 & 0.067 & 0.469 & 0.358
          & 0.407 & 0.083 & 0.513 & 0.093 & 0.472 & 0.314 \\
& L $-9$  & 0.331 & 0.274 & 0.600 & 0.045 & 0.465 & 0.343
          & 0.328 & 0.283 & 0.590 & 0.046 & 0.495 & 0.348 \\
& L $-11$ & 0.394 & 0.272 & 0.598 & 0.105 & 0.491 & 0.372
          & 0.395 & 0.271 & 0.540 & 0.052 & 0.492 & 0.350 \\
\midrule

\Block{4-1}{%
  \parbox[c][3.0cm][c]{0.5cm}{\centering\rotatebox[origin=c]{90}{\textbf{\healthcare}}}
}
& L $-1$  & 0.420 & 0.008 & -0.245 & NA & 0.391 & 0.144
          & 0.198 & 0.009 & 0.306 & NA & 0.471 & 0.246 \\
& L $-7$  & 0.224 & 0.130 & 0.300 & NA & 0.431 & 0.271
          & 0.273 & 0.016 & 0.333 & NA & 0.429 & 0.263 \\
& L $-9$  & 0.345 & 0.071 & 0.277 & NA & 0.411 & 0.276
          & 0.414 & 0.136 & 0.285 & NA & 0.470 & 0.326 \\
& L $-11$ & 0.374 & 0.034 & 0.335 & NA & 0.445 & 0.300
          & 0.455 & 0.116 & 0.325 & NA & 0.424 & 0.330 \\
\midrule

\Block{4-1}{%
  \parbox[c][3.0cm][c]{0.5cm}{\centering\rotatebox[origin=c]{90}{\textbf{\legal}}}
}
& L $-1$  & NA & NA & 0.375 & 0.032 & 0.304 & 0.237
          & NA & NA & 0.479 & 0.100 & 0.339 & 0.306 \\
& L $-7$  & NA & NA & 0.462 & 0.043 & 0.336 & 0.280
          & NA & NA & 0.456 & 0.046 & 0.345 & 0.282 \\
& L $-9$  & NA & NA & 0.467 & 0.113 & 0.365 & 0.315
          & NA & NA & 0.464 & 0.139 & 0.356 & 0.323 \\
& L $-11$ & NA & NA & 0.440 & 0.095 & 0.336 & 0.290
          & NA & NA & 0.447 & 0.092 & 0.321 & 0.287 \\
\midrule

\Block{4-1}{%
  \parbox[c][3.0cm][c]{0.5cm}{\centering\rotatebox[origin=c]{90}{\textbf{\tourism}}}
}
& L $-1$  & -0.105 & 0.468 & NA & 0.176 & NA & 0.180
          & -0.283 & -0.002 & NA & 0.134 & NA & 0.045 \\
& L $-7$  & 0.384 & 0.492 & NA & 0.100 & NA & 0.325
          & 0.396 & 0.512 & NA & 0.052 & NA & 0.320 \\
& L $-9$  & 0.393 & 0.539 & NA & 0.153 & NA & 0.362
          & 0.423 & 0.447 & NA & 0.089 & NA & 0.320 \\
& L $-11$ & 0.360 & 0.524 & NA & -0.057 & NA & 0.276
          & 0.382 & 0.478 & NA & 0.043 & NA & 0.301 \\

\bottomrule
\end{NiceTabular}
\end{adjustbox}

\caption{ALOPE (LoRMA) layer-wise (L) Pearson’s ($r$) scores obtained across domains and language pairs for different adapter configurations.}
\label{tab:lorma_layer_analysis}
\end{table*}

\clearpage

\section{ALOPE: Domain-wise Comparison of Average Performance} \label{app:average_ALOPE_figures}

\begin{figure*}[ht!]
\centering
\begin{minipage}{0.88\linewidth}
\includegraphics[width=\linewidth]{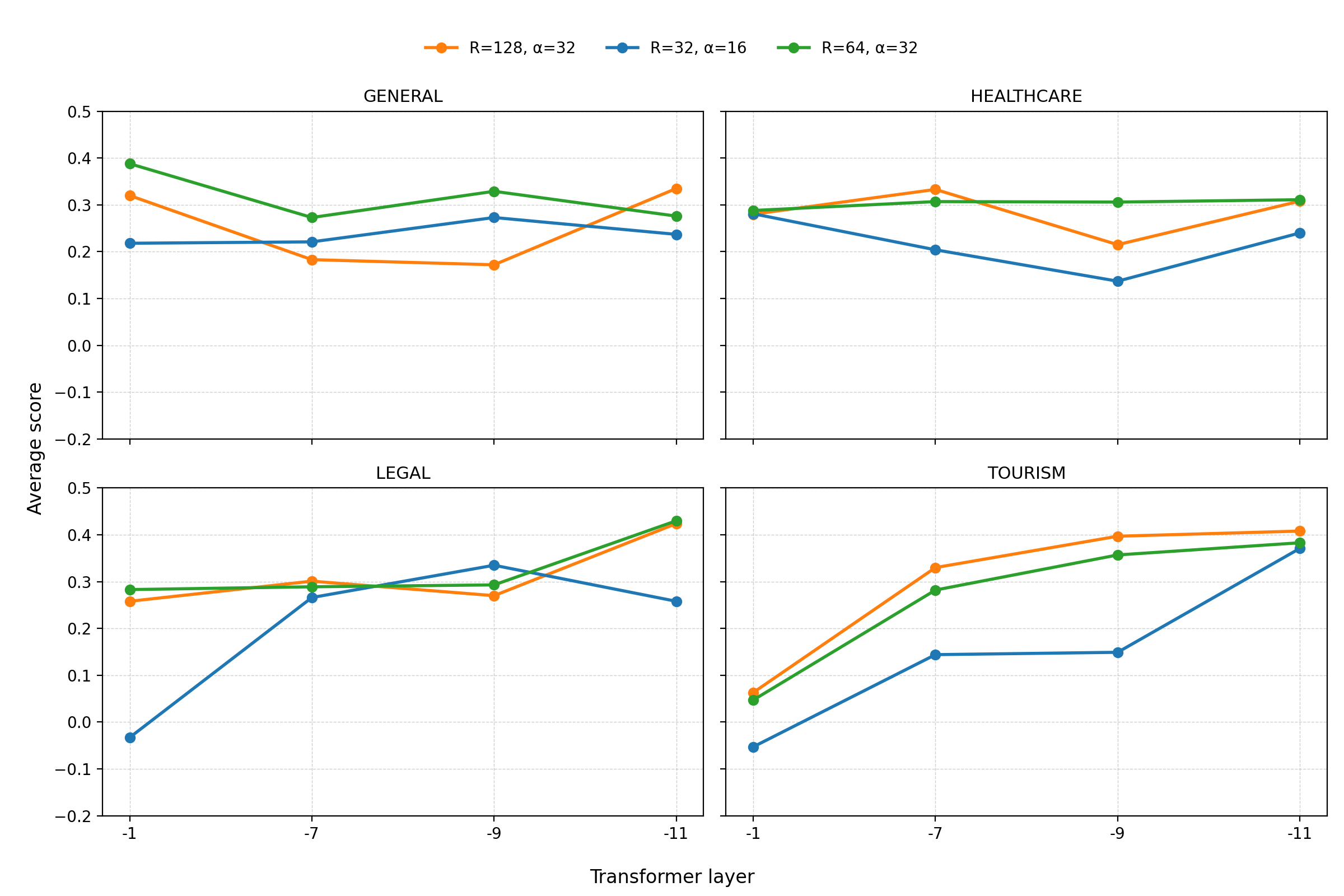}
\caption{ALOPE with LoRA: Average Spearman's ($\rho$) across domains.}
\label{fig:ALOPE}
\end{minipage}
\hfill
\begin{minipage}{0.88\linewidth}
\includegraphics[width=\linewidth]{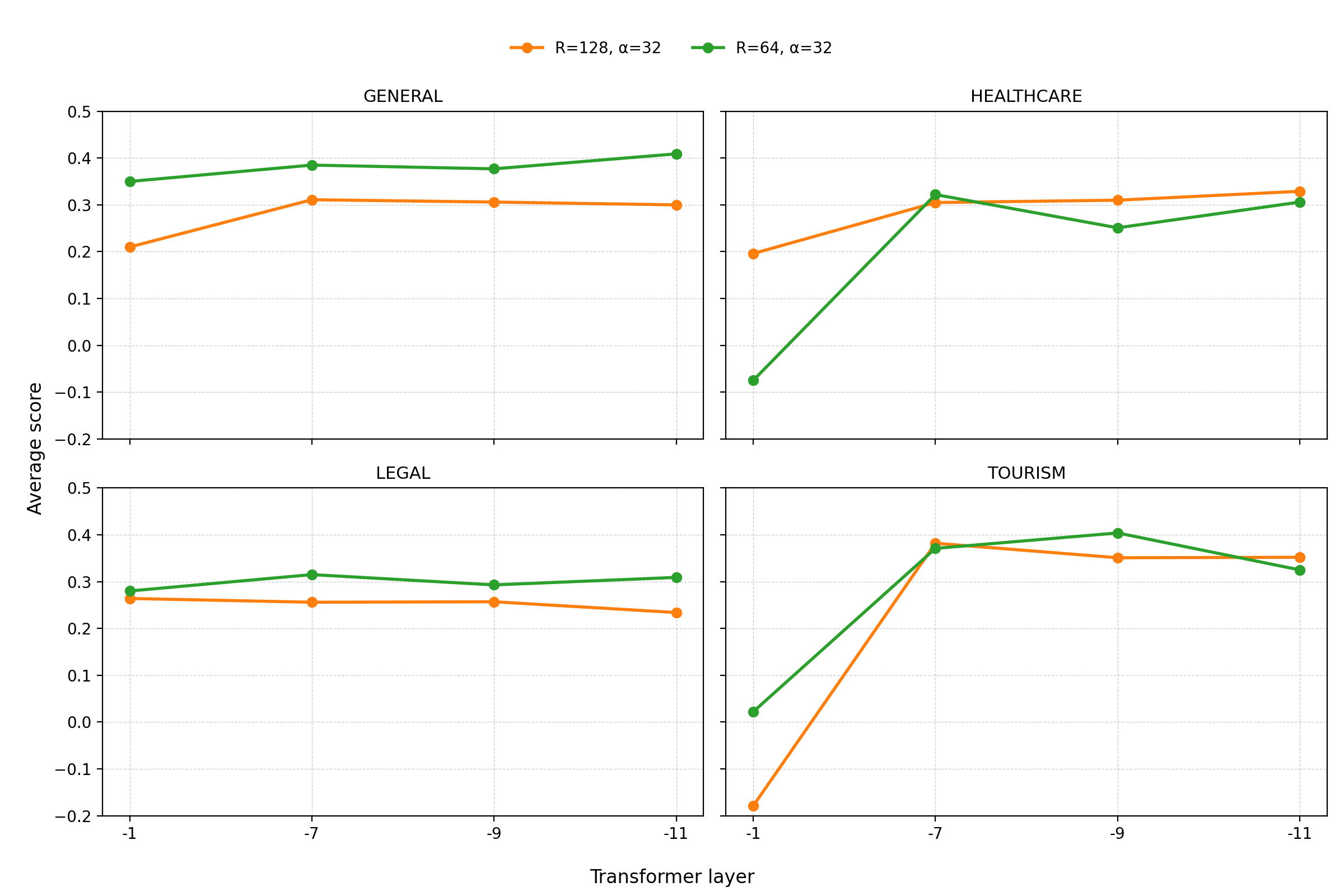}
\caption{ALOPE with LoRMA: Average Spearman's ($\rho$) across domains.}
\label{fig:LORMA}
\end{minipage}
\end{figure*}
\clearpage

\section{Zero-shot Evaluation Results} \label{app:results_zero_prompting}

\begin{table*}[ht]
\centering
\small
\setlength{\tabcolsep}{2.5pt}
\begin{tabular}{llcccccccccccc}
\toprule
\textbf{Domain} & \textbf{Model} & \multicolumn{2}{c}{\textbf{En-Hi}} & \multicolumn{2}{c}{\textbf{En-Mr}} & \multicolumn{2}{c}{\textbf{En-Ta}} & \multicolumn{2}{c}{\textbf{En-Te}} & \multicolumn{2}{c}{\textbf{En-Gu}} & \multicolumn{2}{c}{\textbf{Avg}} \\
\cmidrule(lr){3-4} \cmidrule(lr){5-6} \cmidrule(lr){7-8} \cmidrule(lr){9-10} \cmidrule(lr){11-12} \cmidrule(lr){13-14}
& & $r$ & $\rho$ & $r$ & $\rho$ & $r$ & $\rho$ & $r$ & $\rho$ & $r$ & $\rho$ & $r$ & $\rho$ \\
\midrule
\multirow{6}{*}{\rotatebox{90}{\textbf{\general}}}
& Gemini-1.5-Pro & -0.082 & -0.232 & 0.120 & 0.159 & 0.856 & \textbf{0.848} & 0.191 & 0.073 & 0.774 & \textbf{0.924} & 0.372 & 0.354 \\
& Gemini-2.5-Pro & 0.328 & \textbf{0.424} & 0.747 & \textbf{0.597} & 0.715 & 0.805 & 0.542 & \textbf{0.392} & 0.918 & 0.786 & 0.650 & 0.601 \\
\cmidrule{2-14}
& Qwen3-14B & 0.319 & 0.390 & -0.159 & -0.058 & 0.730 & 0.772 & 0.393 & 0.328 & 0.499 & 0.649 & 0.356 & 0.416 \\
& Gemma3-27B & 0.079 & 0.116 & -0.259 & -0.290 & 0.776 & 0.701 & 0.532 &0.382 & 0.195 & 0.102 & 0.265 & 0.202 \\
& LLaMA-3.2-3B & 0.065 & 0.056 & -0.184 & -0.164 & 0.099 & 0.249 & -0.366 & -0.381 & -0.174 & -0.024 & -0.112 & -0.053 \\
& LLaMA-3.1-8B & -0.156 & -0.284 & -0.512 & -0.465 & 0.587 & 0.437 & 0.265 & 0.115 & 0.662 & 0.812 & 0.169 & 0.123 \\
\midrule
\multirow{6}{*}{\rotatebox{90}{\textbf{\healthcare}}}
& Gemini-1.5-Pro & 0.134 & 0.039 & 0.668 & \textbf{0.814} & -0.068 & 0.056 & NA & NA & 0.369 & 0.346 & 0.276 & 0.294 \\
& Gemini-2.5-Pro & 0.532 & 0.126 & 0.784 & 0.811 & 0.315 & 0.366 & NA & NA & 0.345 & 0.238 & 0.494 & 0.385 \\
\cmidrule{2-14}
& Qwen3-14B & 0.113 & 0.435 & 0.548 & 0.311 & 0.145 & 0.234 & NA & NA & 0.134 & 0.130 & 0.235 & 0.278 \\
& Gemma3-27B & 0.561 & \textbf{0.569} & 0.292 & 0.389 & 0.424 & \textbf{0.603} & NA & NA & 0.523 & \textbf{0.494} & 0.450 & 0.514 \\
& LLaMA-3.2-3B & -0.462 & -0.234 & 0.193 & 0.174 & 0.385 & 0.406 & NA & NA & 0.228 & -0.055 & 0.086 & 0.073 \\
& LLaMA-3.1-8B & 0.001 & -0.071 & -0.196 & -0.087 & -0.180 & 0.100 & NA & NA & -0.143 & -0.395 & -0.13 & -0.113 \\
\midrule
\multirow{6}{*}{\rotatebox{90}{\textbf{\legal}}}
& Gemini-1.5-Pro & NA & NA & NA & NA & 0.899 & \textbf{0.749} & 0.241 & 0.109 & 0.683 & \textbf{0.677} & 0.608 & 0.512 \\
& Gemini-2.5-Pro & NA & NA & NA & NA & 0.353 & 0.475 & 0.043 & 0.055 & 0.607 & 0.457 & 0.334 & 0.329 \\
\cmidrule{2-14}
& Qwen3-14B & NA & NA & NA & NA & 0.343 & 0.418 & -0.689 & -0.676 & 0.240 & 0.265 & -0.035 & 0.002 \\
& Gemma3-27B & NA & NA & NA & NA & -0.206 & -0.056 & -0.035 & -0.045 & 0.260 & 0.201 & 0.006 & 0.033 \\
& LLaMA-3.2-3B & NA & NA & NA & NA & -0.254 & -0.174 & -0.106 & -0.011 & -0.279 & -0.429 & -0.213 & -0.205 \\
& LLaMA-3.1-8B & NA & NA & NA & NA & 0.013 & 0.031 & 0.270 & \textbf{0.287} & 0.045 & -0.105 & 0.109 & 0.071 \\
\midrule
\multirow{6}{*}{\rotatebox{90}{\textbf{\tourism}}}
& Gemini-1.5-Pro & 0.137 & 0.241 & 0.324 & 0.474 & NA & NA & 0.109 & 0.180 & NA & NA & 0.190 & 0.298 \\
& Gemini-2.5-Pro & 0.327 & 0.416 & 0.360 & 0.301 & NA & NA & 0.367 & 0.217 & NA & NA & 0.351 & 0.311 \\
\cmidrule{2-14}
& Qwen3-14B & 0.111 & 0.076 & 0.076 & -0.058 & NA & NA & 0.588 & 0.541 & NA & NA & 0.258 & 0.186 \\
& Gemma3-27B & 0.129 & 0.102 & 0.804 & 0.654 & NA & NA & 0.633 & \textbf{0.636} & NA & NA & 0.522 & 0.464 \\
& LLaMA-3.2-3B & 0.468 & 0.417 & 0.432 & 0.469 & NA & NA & 0.459 & 0.309 & NA & NA & 0.453 & 0.398 \\
& LLaMA-3.1-8B & 0.582 & \textbf{0.613} & 0.839 & \textbf{0.689} & NA & NA & 0.734 & 0.584 & NA & NA & 0.718 & \textbf{0.629} \\
\bottomrule
\end{tabular}
\caption{Zero-shot prompt-only QE performance. Spearman's ($\rho$) and Pearson's ($r$) scores are reported for all language pairs. Best $\rho$ per language pair in \textbf{bold}.`NA' indicates that a language pair is unavailable for that specific domain.}
\label{tab:zero_shot}
\end{table*}
\clearpage

\section{Few-shot with Guidelines Evaluation Results} \label{app:results_few_guideline_prompting}

\begin{table*}[ht]
\centering
\small
\setlength{\tabcolsep}{2.5pt}
\begin{tabular}{llcccccccccccc}
\toprule
\textbf{Domain} & \textbf{Model} & \multicolumn{2}{c}{\textbf{En-Hi}} & \multicolumn{2}{c}{\textbf{En-Mr}} & \multicolumn{2}{c}{\textbf{En-Ta}} & \multicolumn{2}{c}{\textbf{En-Te}} & \multicolumn{2}{c}{\textbf{En-Gu}} & \multicolumn{2}{c}{\textbf{Avg}} \\
\cmidrule(lr){3-4} \cmidrule(lr){5-6} \cmidrule(lr){7-8} \cmidrule(lr){9-10} \cmidrule(lr){11-12} \cmidrule(lr){13-14}
& & $r$ & $\rho$ & $r$ & $\rho$ & $r$ & $\rho$ & $r$ & $\rho$ & $r$ & $\rho$ & $r$ & $\rho$ \\
\midrule
\multirow{6}{*}{\rotatebox{90}{\textbf{\general}}}
& Gemini-1.5-Pro & -0.004 & -0.094 & -0.063 & 0.087 & 0.881 & 0.850 & -0.319 & -0.169 & 0.859 & 0.761 & 0.271 & 0.287 \\
& Gemini-2.5-Pro & 0.625 & \textbf{0.475} & 0.638 & \textbf{0.582} & 0.834 & \textbf{0.886} & 0.255 & 0.238 & 0.926 & 0.776 & 0.656 & 0.591 \\
\cmidrule{2-14}
& Qwen3-14B & 0.247 & 0.279 & -0.159 & -0.058 & 0.720 & 0.783 & 0.498 & \textbf{0.442} & 0.850 & \textbf{0.849} & 0.431 & 0.459 \\
& Gemma3-27B & 0.489 & 0.376 & 0.041 & 0.038 & 0.770 & 0.832 & 0.502 & 0.352 & 0.799 & 0.649 & 0.520 & 0.449 \\
& LLaMA-3.2-3B & -0.637 & -0.527 & -0.101 & -0.251 & 0.479 & 0.329 & 0.150 & 0.186 & -0.332 & -0.200 & -0.088 & -0.093 \\
& LLaMA-3.1-8B & 0.271 & 0.408 & -0.116 & 0.034 & -0.312 & -0.339 & -0.149 & -0.145 & -0.138 & -0.288 & -0.089 & -0.066 \\
\midrule
\multirow{6}{*}{\rotatebox{90}{\textbf{\healthcare}}}
& Gemini-1.5-Pro & 0.170 & 0.415 & 0.614 & 0.669 & -0.237 & -0.040 & NA & NA & 0.604 & 0.100 & 0.288 & 0.286 \\
& Gemini-2.5-Pro & 0.191 & 0.309 & 0.221 & 0.024 & -0.302 & -0.089 & NA & NA & 0.584 & 0.084 & 0.174 & 0.082 \\
\cmidrule{2-14}
& Qwen3-14B & 0.176 & 0.225 & 0.886 & \textbf{0.884} & -0.144 & -0.240 & NA & NA & -0.357 & -0.389 & 0.140 & 0.120 \\
& Gemma3-27B & 0.255 & 0.345 & 0.193 & 0.174 & 0.076 & 0.123 & NA & NA & 0.225 & 0.109 & 0.187 & 0.188 \\
& LLaMA-3.2-3B & 0.012 & 0.058 & -0.499 & -0.285 & -0.476 & -0.284 & NA & NA & -0.381 & -0.479 & -0.336 & -0.248 \\
& LLaMA-3.1-8B & 0.204 & \textbf{0.447} & -0.541 & -0.458 & 0.257 & \textbf{0.411} & NA & NA & 0.612 & \textbf{0.211} & 0.133 & 0.153 \\
\midrule
\multirow{6}{*}{\rotatebox{90}{\textbf{\legal}}}
& Gemini-1.5-Pro & NA & NA & NA & NA & 0.867 & \textbf{0.717} & 0.446 & \textbf{0.530} & -0.078 & 0.072 & 0.412 & 0.440 \\
& Gemini-2.5-Pro & NA & NA & NA & NA & 0.615 & 0.532 & 0.375 & 0.324 & 0.849 & 0.699 & 0.613 & 0.518 \\
\cmidrule{2-14}
& Qwen3-14B & NA & NA & NA & NA & 0.343 & 0.418 & -0.689 & -0.676 & 0.240 & 0.265 & -0.035 & 0.002 \\
& Gemma3-27B & NA & NA & NA & NA & 0.249 & 0.099 & 0.195 & 0.102 & 0.699 & \textbf{0.727} & 0.381 & 0.309 \\
& LLaMA-3.2-3B & NA & NA & NA & NA & -0.333 & -0.477 & 0.378 & 0.528 & 0.293 & 0.304 & 0.113 & 0.118 \\
& LLaMA-3.1-8B & NA & NA & NA & NA & 0.254 & 0.104 & 0.121 & 0.030 & -0.063 & 0.087 & 0.104 & 0.074 \\
\midrule
\multirow{6}{*}{\rotatebox{90}{\textbf{\tourism}}}
& Gemini-1.5-Pro & 0.447 & \textbf{0.502} & 0.630 & 0.480 & NA & NA & 0.174 & 0.308 & NA & NA & 0.417 & 0.430 \\
& Gemini-2.5-Pro & 0.379 & 0.385 & 0.755 & 0.679 & NA & NA & 0.327 & 0.472 & NA & NA & 0.487 & 0.512 \\
\cmidrule{2-14}
& Qwen3-14B & 0.059 & 0.060 & 0.371 & 0.236 & NA & NA & 0.433 & \textbf{0.583} & NA & NA & 0.288 & 0.293 \\
& Gemma3-27B & 0.226 & 0.078 & 0.852 & \textbf{0.702} & NA & NA & 0.002 & 0.051 & NA & NA & 0.360 & 0.277 \\
& LLaMA-3.2-3B & -0.471 & -0.459 & -0.248 & -0.098 & NA & NA & -0.118 & 0.032 & NA & NA & -0.279 & -0.175 \\
& LLaMA-3.1-8B & 0.008 & 0.158 & -0.208 & -0.058 & NA & NA & -0.068 & -0.218 & NA & NA & -0.089 & -0.039 \\
\bottomrule
\end{tabular}
\caption{Few-shot with guidelines QE performance. Spearman's ($\rho$) and Pearson's ($r$) scores are reported for all language pairs. Best $\rho$ per language pair in \textbf{bold}. `NA' indicates that a language pair is unavailable for that specific domain.}
\label{tab:few_shot_guidelines}
\end{table*}

\clearpage

\section{Few-shot without Guidelines Evaluation Results} \label{app:results_few_without_guideline_prompting}

\begin{table*}[ht]
\centering
\small
\setlength{\tabcolsep}{2.5pt}
\begin{tabular}{llcccccccccccc}
\toprule
\textbf{Domain} & \textbf{Model} & \multicolumn{2}{c}{\textbf{En-Hi}} & \multicolumn{2}{c}{\textbf{En-Mr}} & \multicolumn{2}{c}{\textbf{En-Ta}} & \multicolumn{2}{c}{\textbf{En-Te}} & \multicolumn{2}{c}{\textbf{En-Gu}} & \multicolumn{2}{c}{\textbf{Avg}} \\
\cmidrule(lr){3-4} \cmidrule(lr){5-6} \cmidrule(lr){7-8} \cmidrule(lr){9-10} \cmidrule(lr){11-12} \cmidrule(lr){13-14}
& & $r$ & $\rho$ & $r$ & $\rho$ & $r$ & $\rho$ & $r$ & $\rho$ & $r$ & $\rho$ & $r$ & $\rho$ \\
\midrule
\multirow{6}{*}{\rotatebox{90}{\textbf{\general}}}
& Gemini-1.5-Pro & 0.107 & 0.022 & -0.213 & -0.103 & 0.744 & 0.743 & -0.035 & -0.185 & 0.941 & 0.791 & 0.309 & 0.254 \\
& Gemini-2.5-Pro & 0.713 & \textbf{0.563} & 0.447 & 0.314 & 0.888 & \textbf{0.940} & -0.095 & -0.031 & 0.939 & \textbf{0.860} & 0.578 & 0.529 \\
\cmidrule{2-14}
& Qwen3-14B & -0.128 & -0.040 & -0.235 & -0.234 & 0.761 & 0.867 & 0.319 & 0.235 & 0.658 & 0.752 & 0.275 & 0.316 \\
& Gemma3-27B & 0.177 & 0.234 & 0.489 & \textbf{0.375} & 0.519 & 0.450 & 0.636 & \textbf{0.486} & 0.765 & 0.627 & 0.517 & 0.434 \\
& LLaMA-3.2-3B & 0.497 & 0.418 & -0.101 & 0.014 & 0.393 & 0.477 & 0.223 & 0.188 & 0.078 & 0.228 & 0.218 & 0.265 \\
& LLaMA-3.1-8B & 0.234 & 0.084 & 0.016 & 0.109 & -0.332 & -0.482 & -0.180 & -0.177 & 0.097 & 0.005 & -0.033 & -0.092 \\
\midrule
\multirow{6}{*}{\rotatebox{90}{\textbf{\healthcare}}}
& Gemini-1.5-Pro & 0.568 & 0.307 & 0.852 & 0.552 & -0.027 & 0.168 & NA & NA & 0.608 & \textbf{0.458} & 0.500 & 0.371 \\
& Gemini-2.5-Pro & 0.670 & 0.585 & 0.712 & 0.786 & 0.060 & -0.086 & NA & NA & 0.388 & 0.393 & 0.458 & 0.420 \\
\cmidrule{2-14}
& Qwen3-14B & 0.261 & 0.413 & 0.865 & \textbf{0.884} & -0.117 & -0.075 & NA & NA & -0.049 & -0.143 & 0.240 & 0.270 \\
& Gemma3-27B & 0.083 & 0.187 & 0.097 & 0.114 & 0.132 & 0.275 & NA & NA & 0.216 & 0.174 & 0.132 & 0.188 \\
& LLaMA-3.2-3B & 0.513 & \textbf{0.611} & 0.123 & 0.035 & 0.648 & \textbf{0.422} & NA & NA & 0.139 & 0.071 & 0.356 & 0.285 \\
& LLaMA-3.1-8B & 0.519 & 0.406 & -0.244 & 0.289 & -0.101 & 0.250 & NA & NA & 0.283 & 0.398 & 0.114 & 0.336 \\
\midrule
\multirow{6}{*}{\rotatebox{90}{\textbf{\legal}}}
& Gemini-1.5-Pro & NA & NA & NA & NA & 0.571 & 0.475 & 0.355 & 0.230 & 0.170 & 0.061 & 0.365 & 0.255 \\
& Gemini-2.5-Pro & NA & NA & NA & NA & -0.342 & -0.492 & -0.262 & -0.112 & 0.625 & \textbf{0.475} & 0.007 & -0.043 \\
\cmidrule{2-14}
& Qwen3-14B & NA & NA & NA & NA & 0.350 & 0.467 & 0.365 & 0.215 & 0.332 & 0.473 & 0.349 & 0.385 \\
& Gemma3-27B & NA & NA & NA & NA & 0.277 & 0.311 & 0.828 & \textbf{0.687} & 0.304 & 0.211 & 0.470 & 0.403 \\
& LLaMA-3.2-3B & NA & NA & NA & NA & -0.471 & -0.459 & -0.248 & -0.098 & -0.118 & 0.032 & -0.279 & -0.175 \\
& LLaMA-3.1-8B & NA & NA & NA & NA & 0.887 & \textbf{0.737} & -0.438 & -0.335 & 0.094 & -0.026 & 0.181 & 0.125 \\
\midrule
\multirow{6}{*}{\rotatebox{90}{\textbf{\tourism}}}
& Gemini-1.5-Pro & 0.142 & 0.193 & 0.835 & 0.685 & NA & NA & -0.126 & -0.208 & NA & NA & 0.284 & 0.223 \\
& Gemini-2.5-Pro & 0.359 & 0.509 & 0.501 & 0.406 & NA & NA & 0.422 & 0.397 & NA & NA & 0.427 & 0.437 \\
\cmidrule{2-14}
& Qwen3-14B & 0.350 & 0.467 & 0.365 & 0.215 & NA & NA & 0.332 & \textbf{0.473} & NA & NA & 0.349 & 0.485 \\
& Gemma3-27B & 0.277 & 0.311 & 0.828 & \textbf{0.687} & NA & NA & 0.304 & 0.211 & NA & NA & 0.470 & 0.403 \\
& LLaMA-3.2-3B & -0.471 & -0.459 & -0.248 & -0.098 & NA & NA & -0.118 & 0.032 & NA & NA & -0.279 & 0.175 \\
& LLaMA-3.1-8B & 0.887 & \textbf{0.737} & -0.438 & -0.335 & NA & NA & 0.094 & -0.026 & NA & NA & 0.181 & 0.125 \\
\bottomrule
\end{tabular}
\caption{Few-shot without guidelines QE performance. Spearman's ($\rho$) and Pearson's ($r$) scores are reported for all language pairs. Best $\rho$ per language pair in \textbf{bold}. `NA' indicates that a language pair is unavailable for that specific domain.}
\label{tab:few_shot_no_guidelines}
\end{table*}

\clearpage

\section{Comparison of ALOPE and Prompt-only Approaches}  \label{app:domain_prompt_vs_alope}

\begin{table*}[ht]
\centering
\small
\setlength{\tabcolsep}{4pt}

\begin{NiceTabular}{l|l|ccccc|c}
\toprule
\textbf{Domain} & \textbf{Prompt setting} &
\textbf{En-Hi} & \textbf{En-Mr} & \textbf{En-Ta} & \textbf{En-Te} & \textbf{En-Gu} & \textbf{Avg} \\
\midrule

\Block{5-1}{\textbf{\general}}
& Zero-shot & 0.056 & -0.164 & 0.249 & -0.381 & -0.024 & -0.053 \\
& Few-shot + Guidelines & -0.527 & -0.251 & 0.329 & 0.186 & -0.200 & -0.093 \\
& Few-shot (No Guidelines) & 0.418 & 0.014 & 0.477 & 0.188 & 0.228 & 0.265 \\
& ALOPE (LoRA) & \textbf{0.476} & 0.157 & \textbf{0.610} & \textbf{0.292} & \textbf{0.485} & $\dagger$0.404 \\
& ALOPE (LoRMA) & 0.392 & \textbf{0.314} & 0.507 & 0.168 & 0.421 & 0.360 \\
\midrule

\Block{5-1}{\textbf{\healthcare}}
& Zero-shot & -0.234 & 0.174 & 0.406 & NA & -0.055 & 0.073 \\
& Few-shot + Guidelines & 0.058 & -0.285 & -0.284 & NA & -0.479 & -0.248 \\
& Few-shot (No Guidelines) & \textbf{0.611} & 0.035 & 0.422 & NA & 0.071 & 0.285 \\
& ALOPE (LoRA) & 0.520 & \textbf{0.192} & \textbf{0.415} & NA & \textbf{0.532} & $\dagger$0.415 \\
& ALOPE (LoRMA) & 0.365 & 0.131 & 0.413 & NA & 0.512 & 0.355 \\
\midrule

\Block{5-1}{\textbf{\legal}}
& Zero-shot & NA & NA & -0.174 & -0.011 & -0.429 & -0.205 \\
& Few-shot + Guidelines & NA & NA & -0.477 & \textbf{0.528} & 0.304 & 0.118 \\
& Few-shot (No Guidelines) & NA & NA & -0.459 & -0.098 & 0.032 & -0.175 \\
& ALOPE (LoRA) & NA & NA & \textbf{0.581} & 0.267 & \textbf{0.445} & $\dagger$0.431 \\
& ALOPE (LoRMA) & NA & NA & 0.515 & 0.125 & 0.216 & 0.265 \\
\midrule

\Block{5-1}{\textbf{\tourism}}
& Zero-shot & 0.417 & 0.469 & NA & \textbf{0.309} & NA & 0.398 \\
& Few-shot + Guidelines & -0.459 & -0.098 & NA & 0.032 & NA & -0.175 \\
& Few-shot (No Guidelines) & -0.459 & -0.098 & NA & 0.032 & NA & -0.175 \\
& ALOPE (LoRA) & 0.350 & \textbf{0.670} & NA & 0.220 & NA & $\dagger$0.413 \\
& ALOPE (LoRMA) & \textbf{0.465} & 0.532 & NA & 0.227 & NA & 0.408 \\

\bottomrule
\end{NiceTabular}

\caption{Comparison of Spearman’s correlation ($\rho$) achieved by LLaMA-3.2-3B across Prompt-only approaches and ALOPE. The bolded values represent the highest Spearman scores obtained for each language pair in each domain. The ($\dagger$) represents the highest average obtained in each domain. `NA' indicates that a language pair is unavailable for that specific domain.}
\label{tab:domain_prompt_vs_alope}
\end{table*}

\end{document}